\pdfoutput=1

\documentclass[11pt]{article}

\usepackage{acl}

\usepackage{amsmath}
\usepackage{times}
\usepackage{latexsym}
\usepackage{hyperref}
\usepackage[noabbrev,capitalize,nameinlink]{cleveref}
\usepackage{xcolor, soul, colortbl}
\usepackage{array, multirow, graphicx}
\usepackage{adjustbox}
\usepackage[T1]{fontenc}
\usepackage{booktabs}
\usepackage{tikz}
\usepackage{tikz-dependency}
\usepackage{lscape, float} 
\usepackage{changes}
\usepackage[shortlabels]{enumitem}
\usepackage{fontawesome}

\usepackage[utf8]{inputenc}

\usepackage{microtype}
\definecolor{Gray}{gray}{0.9}
\definecolor{LightCyan}{rgb}{0.88,1,1}
\newcolumntype{?}{!{\vrule width 1pt}}
\newcolumntype{g}{>{\columncolor{LightCyan}}r}

\newcommand{\bertb}{BERT\textsubscript{base}}

\newcommand{\sk}[1]{\colorbox{pink}{$\textrm{[#1]}_\textrm{\tiny SKILL}$}}
\newcommand{\kn}[1]{\colorbox{yellow}{$\textrm{[#1]}_\textrm{\tiny KNOWLEDGE}$}}
\newcommand{\guide}[1]{\noindent\fbox{%
\parbox{\textwidth}{%
#1}%
}}
\newcommand{\bjo}{\textsc{Big}}
\newcommand{\hou}{\textsc{House}}
\newcommand{\tech}{\textsc{Tech}}
\newcommand{\std}[2]{{#1}{\footnotesize$\pm${#2}}}
\newcommand*\circled[1]{\tikz[baseline=(char.base)]{
            \node[shape=circle,draw,inner sep=.6pt] (char) {#1};}}
\usepackage{amssymb}
\usepackage{pifont}
\newcommand{\cmark}{\ding{51}}%
\newcommand{\xmark}{\ding{55}}%

\title{\textsc{{S}kill{S}pan}: Hard and Soft Skill Extraction from {E}nglish Job Postings}

 \author{Mike Zhang\textsuperscript{*}\textsuperscript{$\diamondsuit$} \hspace{5pt} Kristian Nørgaard Jensen\textsuperscript{*}\textsuperscript{$\diamondsuit$} \hspace{5pt} Sif Dam Sonniks\textsuperscript{$\diamondsuit$} \hspace{5pt} Barbara Plank\textsuperscript{$\diamondsuit$}\textsuperscript{$\clubsuit$} \\
  \textsuperscript{$\diamondsuit$}Department of Computer Science, IT University of Copenhagen, Denmark\\
  \textsuperscript{$\clubsuit$}Center for Information and Language Processing (CIS), LMU Munich, Germany\\
{\tt \{mikz, krnj, sifs\}@itu.dk } \hspace{2em} {\tt bplank@cis.uni-muenchen.de}}

\begin{document}
\maketitle
\begingroup\renewcommand\thefootnote{*}
\footnotetext{Equal contribution.}
\endgroup
\begin{abstract}
Skill Extraction (SE) is an important and widely-studied task useful to gain insights into labor market dynamics. However, there is a lacuna of datasets and annotation guidelines; available datasets are few and contain crowd-sourced labels on the span-level or labels from a predefined skill inventory.
To address this gap, we introduce \textsc{SkillSpan}, a novel SE dataset consisting of 14.5K sentences and over 12.5K annotated spans. We release its respective guidelines created over three different sources annotated for hard \emph{and} soft skills by domain experts. We introduce a BERT baseline~\cite{devlin2019bert}. To improve upon this baseline, we experiment with language models that are optimized for long spans~\cite{joshi2020spanbert, beltagy2020longformer}, continuous pre-training on the job posting domain~\cite{han-eisenstein-2019-unsupervised, gururangan2020don}, and multi-task learning~\cite{caruana1997multitask}. Our results show that the domain-adapted models significantly outperform their non-adapted counterparts, and single-task outperforms multi-task learning. 
\end{abstract}

\section{Introduction}

Job markets are under constant development---often due to developments in technology, migration, and digitization---so are the skill sets required. Consequentially, job vacancy data is emerging on a variety of platforms in big quantities and can provide insights on labor market skill demands or aid job matching~\cite{balog2012expertise}. SE is to extract the competences necessary from unstructured text. 

Previous work in SE shows promising progress, but is halted by a lack of available datasets and annotation guidelines. Two out of 14 studies release their dataset, which limit themselves to crowd-sourced labels~\cite{sayfullina2018learning} or annotations from a predefined list of skills on the document-level~\cite{bhola-etal-2020-retrieving}. Additionally, none of the 14 previously mentioned studies release their annotation guidelines, which obscures the meaning of a competence. Job markets change, as do the skills in, e.g., the European Skills, Competences, Qualifications and Occupations~\citep[ESCO;\ ][]{le2014esco} taxonomy (\cref{skillknowledge}). Hence, it is important to cover for possible emerging skills.

\begin{figure}
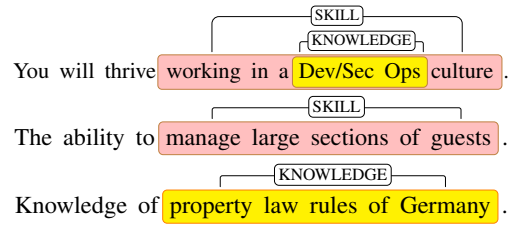

\centering
\resizebox{.9\linewidth}{!}{
\begin{dependency}[edge slant=0pt, edge vertical padding=3pt, edge style={-}]
    \begin{deptext}
    You \& will \& thrive \& working \& in \& a \& Dev/Sec \& Ops \& culture \& . \\
    \end{deptext}
    \wordgroup[group style={fill=pink, draw=brown, inner sep=.3ex}]{1}{4}{9}{sk}
    \wordgroup[group style={fill=yellow, draw=brown, inner sep=.1ex}]{1}{7}{8}{kn}
    \depedge[edge height=3.5ex]{4}{9}{SKILL}
    \depedge[edge height=1ex, edge horizontal padding=-18pt, edge end x offset=10pt]{7}{8}{KNOWLEDGE}
\end{dependency}}
\resizebox{.9\linewidth}{!}{
\begin{dependency}[edge slant=0pt, edge vertical padding=2pt, edge style={-}]
    \begin{deptext}
    The \& ability \& to \& manage \& large \& sections \& of \& guests  \& .\\
    \end{deptext}
    \wordgroup[group style={fill=pink, draw=brown, inner sep=.1ex}]{1}{4}{8}{sk}
    \depedge[edge height=1ex]{4}{8}{SKILL}
\end{dependency}}
\resizebox{.9\linewidth}{!}{
\begin{dependency}[edge slant=0pt, edge vertical padding=2pt, edge style={-}]
    \begin{deptext}
    Knowledge \& of \& property \& law \& rules \& of \& Germany \& .\\
    \end{deptext}
    \wordgroup[group style={fill=yellow, draw=orange, inner sep=.1ex}]{1}{3}{7}{kn}
    \depedge[edge height=1ex]{3}{7}{KNOWLEDGE}
\end{dependency}}
\looseness=-1
    \caption{\textbf{Examples of Skills \& Knowledge Components.} Annotated samples of passages in varying job postings. More details are given in \cref{sec:ann}.}
    \label{fig:doccano}
\end{figure}

We propose \textsc{SkillSpan}, a novel SE dataset annotated at the span-level for \emph{skill} and \emph{knowledge} components (SKCs) in job postings (JPs). As illustrated in~\cref{fig:doccano}, SKCs can be nested inside skills. \textsc{SkillSpan} allows for extracting possibly undiscovered competences and to diminish the lack of coverage of predefined skill inventories.

Our analysis~(\cref{fig:violin}) shows that SKCs contain on average longer sequences than typical Named Entity Recognition (NER) tasks. Albeit we additionally study models optimized for long spans~\cite{joshi2020spanbert, beltagy2020longformer}, some underperform. Overall, we find specialized domain BERT models~\cite{alsentzer-etal-2019-publicly, lee2020biobert, gururangan2020don, nguyen-etal-2020-bertweet} perform better than their non-adapted counterparts. We explore the benefits of domain-adaptive pre-training on the JP domain~\cite{han-eisenstein-2019-unsupervised, gururangan2020don}. Last, given the examples from~\cref{fig:doccano}, we formulate the task as both as a sequence labeling and a multi-task learning (MTL) problem, i.e., training on both skill and knowledge components jointly~\cite{caruana1997multitask}. 

\begin{table*}[t]
\centering
    \resizebox{\linewidth}{!}{
    \begin{tabular}{l|lllllll}
    \toprule
& \textbf{Annotations} & \textbf{Approach} & \textbf{Size} & \textbf{Skill Type} & \textbf{(Baseline) Model(s)} & \faPencil & \faBook\\
\midrule
 \citet{kivimaki-etal-2013-graph}   & Document-level        &  Automatic        & N/A               & Hard &  LogEnt., TF-IDF, LSA, LDA         & \xmark & \xmark \\
 \citet{zhao2015skill}              & Sentence-level        &  Automatic        & N/A               & Hard & Word2Vec                           & \xmark & \xmark\\
 \citet{javed2017large}             & Span-level           &  Skill Inventory  & N/A               & Both & Word2Vec                            & \xmark & \xmark\\
 \citet{jia2018representation}      & Span-level       &  Automatic        & 21,158 JPs*       & Hard & LSTM                                    & \xmark & \xmark\\
 \citet{sayfullina2018learning}     & Span-level        &  Crowdsourcing    & 4,863 Sent.       & Soft & CNN, LSTM                            & \cmark & \xmark\\
 \citet{smith2019syntax}            & Span-level           &  Manual           & 100 JPs           & Hard & Pattern Matching                     & \xmark & \xmark\\
 \citet{gugnani2020implicit}        & Span-level           &  Domain Experts   & $\sim$200 JPs     & Hard & Word2Vec, Doc2Vec                   & \xmark & \xmark\\
 \citet{li2020deep}                 & Document-level        &  Proprietary      & N/A               & Hard & FastText                           & \xmark & \xmark\\
 \citet{shi2020salience}            & Span-level           &  Proprietary      & N/A               & Hard & FastText, USE, BERT                 & \xmark & \xmark\\
 \citet{tamburri2020dataops}        & Sentence-level        &  Domain Experts   & $\sim$3,000 Sent. & Both & BERT                               & \xmark & \xmark\\
 \citet{chernova2020occupational}   & Span-level            &  Manual           & 100 JPs           & Both & FinBERT                            & \xmark & \xmark\\
 \citet{bhola-etal-2020-retrieving} & Document-level        &  Skill Inventory  & 20,298 JPs*       & Hard & BERT                               & \cmark & \xmark\\
 \citet{smith2021skill}             & Span-level           &  Manual           & 100 JPs           & Hard & Pattern Match., Word2Vec            & \xmark & \xmark\\
 \citet{liu2021learning}            & Document-level        &  Crowdsourcing    & N/A               & Hard & GNN                                & \xmark & \xmark\\
 \midrule
 \textbf{This work}                 & Span-level    &  Domain Experts   & 391 JPs           & Both & (Domain-adapted) BERT           & \cmark  & \cmark \\
    \bottomrule
    \end{tabular}}
    \caption{\textbf{Contributions of Related Work}. We list the recent works of Skill Extraction. Note that (*) indicates labels that are automatically inferred from some source (e.g., a predefined skill inventory) and not \emph{manually} annotated. With respect to the annotation approach, ``Manual'' indicates uncertainty whether they used domain experts or not. Also note that many works do not release their dataset with annotations (\faPencil)  nor guidelines (\faBook). The list is inspired by \citet{khaouja2021survey}.}
    \label{tab:relwork}
\end{table*}

\paragraph{Contributions} In this paper: 
\circled{1} We release \textsc{SkillSpan}, a novel skill extraction dataset, 
with annotation guidelines, and our open-source code.\footnote{\url{https://github.com/kris927b/SkillSpan}} \circled{2} We present strong baselines for the task including a new SpanBERT~\cite{joshi2020spanbert} trained from scratch, and domain-adapted variants~\cite{gururangan2020don}, which we will release on the \texttt{HuggingFace} platform~\cite{wolf-etal-2020-transformers}. To the best of our knowledge, we are the first to investigate the extraction of skills and knowledge from job postings with state-of-the-art language models.
\circled{3} We give an analysis on single-task versus multi-task learning in the context of skill extraction, and show that for this particular task single-task learning outperforms multi-task learning. 

\section{Related Work}

There is a pool of prior work relating to SE. We summarize it in~\cref{tab:relwork}, depicting state-of-the-art approaches, level of annotations, what kind of competences are annotated, the modeling approaches, the size of the dataset (if available), type of skills annotated for, baseline models, and whether they release their annotations and guidelines.

As can be seen in~\cref{tab:relwork}, many works do \textit{not release their data} (apart from~\citealp{sayfullina2018learning} and~\citealp{bhola-etal-2020-retrieving}) and \textbf{none release their annotation guidelines}. In addition, none of the previous studies approach SE as a span-level extraction task with state-of-the-art language models, nor did they release a dataset of this magnitude with manually annotated (long) spans of competences by domain experts.

Although \citet{sayfullina2018learning} annotated on the span-level (thus being useful for SE) and release their data, they instead explored several approaches to \emph{Skill Classification}. To create the data, they extracted all text snippets containing \emph{one} soft skill from a predetermined list. Crowdworkers then annotated the highlighted skill whether it was a soft skill referring to the candidate or not. They show that an LSTM~\cite{hochreiter1997long} performs best on classifying the skill in the sentence. In our work, we annotated a dataset three times their size (\cref{tab:num_post}) for both hard \emph{and} soft skills. In addition, we also extract the specific skills from the sentence.

\citet{tamburri2020dataops} classifies sentences that contain skills in the JP. The authors manually labeled their dataset with domain experts. They annotated whether a sentence contains a skill or not. Once the sentence is identified as containing a skill, the skill cited within is extracted. In contrast, we directly annotate for the span within the sentence.

\citet{bhola-etal-2020-retrieving} cast the task of skill extraction as a multi-label skill classification at the document-level. There is a predefined set of unique skills given the job descriptions and they predict multiple skills that are connected to a given job description using BERT~\cite{devlin2019bert}. In addition, they experiment with several additional layers for better prediction performance. We instead explore domain-adaptive pre-training for SE.

The work closest to ours is by \citet{chernova2020occupational}, who approach the task similarly with span-level annotations (including longer spans) but approach this for the Finnish language. It is unclear whether they annotated by domain experts. Also, neither the data nor the annotation guidelines are released. For a comprehensive overview with respect to SE, we refer to~\citet{khaouja2021survey}.

\section{Skill \& Knowledge Definition}\label{skillknowledge}
There is an abundance of competences and there have been large efforts to categorize them. For example, the The International Standard Classification of Occupations~\citep[ISCO;\ ][]{elias1997occupational} is one of the main international classifications of occupations and skills. It belongs to the international family of economic and social classifications. Another example, the European Skills, Competences, Qualifications and Occupations~\citep[ESCO;\ ][]{le2014esco} taxonomy is the European standard terminology linking skills and competences and qualifications to occupations and derived from ISCO. The ESCO taxonomy mentions three categories of competences: \emph{Skill}, \emph{knowledge}, and \emph{attitudes}. ESCO defines knowledge as follows:
\begin{quote}
    ``Knowledge means the outcome of the assimilation of information through learning. Knowledge is the body of facts, principles, theories and practices that is related to a field of work or study.''~\footnote{\url{https://ec.europa.eu/esco/portal/escopedia/Knowledge}}
\end{quote}
For example, a person can acquire the Python programming language through learning. This is denoted as a \emph{knowledge} component and can be considered a \emph{hard skill}. However, one also needs to be able to apply the knowledge component to a certain task. This is known as a \emph{skill} component. ESCO formulates it as:
\begin{quote}
    ``Skill means the ability to apply knowledge and use know-how to complete tasks and solve problems.''~\footnote{\url{https://ec.europa.eu/esco/portal/escopedia/Skill}}
\end{quote}
 
\noindent
In ESCO, the \emph{soft skills} are referred to as \emph{attitudes}. ESCO considers attitudes as skill components:

\begin{quote}
    ``The ability to use knowledge, skills and personal, social and/or methodological abilities, in work or study situations and professional and personal development.''~\footnote{\url{http://data.europa.eu/esco/skill/A}}
\end{quote}

\noindent
To sum up, hard skills are usually referred to as \emph{knowledge} components, and applying these hard skills to something is considered a \emph{skill} component. Then, soft skills are referred to as \emph{attitudes}, these are part of skill components. There has been no work, to the best of our knowledge, in annotating skill and knowledge components in JPs.

\section{\textsc{SkillSpan} Dataset}\label{sec:ann}

\begin{table}
\centering
\resizebox{\linewidth}{!}{
    \begin{tabular}{l|lrrr||g}
    \toprule
& $\downarrow$ \textbf{Statistics}, \textbf{Src.} $\rightarrow$ & \textbf{\bjo{}} & \textbf{\hou{}} & \textbf{\tech{}} & \textbf{Total}\\
\midrule
\multirow{6}{*}{\rotatebox[origin=c]{90}{\textbf{Train}}}
   & \textbf{\# Posts}                  & 60        & 60        & 80        & 200       \\
   & \textbf{\# Sentences}              & 1,036     & 1,674     & 3,156     & 5,866     \\
   & \textbf{\# Tokens}                 & 29,064    & 36,995    & 56,549    & 122,608   \\
   & \textbf{\# Skill Spans}            & 1,086     & 984       & 1,237     & 3,307     \\
   & \textbf{\# Knowledge Spans}        & 439       & 781       & 2,188     & 3,408     \\
   & \textbf{\# Overlapping Spans}      & 45        & 29        & 135       & 209       \\
   \midrule
\multirow{6}{*}{\rotatebox[origin=c]{90}{\textbf{Development}}}
   & \textbf{\# Posts}                  & 30        & 30        & 30        & 90     \\
   & \textbf{\# Sentences}              & 783       & 1,022     & 2,187     & 3,992  \\
   & \textbf{\# Tokens}                 & 11,762    & 19,173    & 21,149    & 52,084 \\
   & \textbf{\# Skill Spans}            & 469       & 525       & 545       & 1,539  \\
   & \textbf{\# Knowledge Spans}        & 126       & 287       & 806       & 1,219  \\
   & \textbf{\# Overlapping Spans}      & 12        & 17        & 32        & 61     \\
   \midrule
\multirow{6}{*}{\rotatebox[origin=c]{90}{\textbf{Test}}}
   & \textbf{\# Posts}                  & 36        & 33        & 32        & 101    \\
   & \textbf{\# Sentences}              & 1,112     & 1,216     & 2,352     & 4,680  \\
   & \textbf{\# Tokens}                 & 14,720    & 21,923    & 20,885    & 57,528 \\
   & \textbf{\# Skill Spans}            & 634       & 637       & 459       & 1,730  \\
   & \textbf{\# Knowledge Spans}        & 242       & 350       & 834       & 1,426  \\
   & \textbf{\# Overlapping Spans}      & 12        & 8         & 9         & 29     \\
  \midrule
  \midrule
  \rowcolor{LightCyan}
   & \textbf{\# Posts}                  & 126       & 123       & 142       & 391       \\
  \rowcolor{LightCyan}
   & \textbf{\# Sentences}              & 2,931     & 3,912     & 7,695     & 14,538    \\
    \rowcolor{LightCyan}
   & \textbf{\# Tokens}                 & 55,546    & 78,091    & 98,583    & 232,220   \\
    \rowcolor{LightCyan}
   & \textbf{\# Skill Spans}            & 2,189     & 2,146     &  2,241    & 6,576     \\
   \rowcolor{LightCyan}
   & \textbf{\# Knowledge Spans}        & 807       & 1,418     &  3,828    & 6,053     \\
   \rowcolor{LightCyan}
   \multirow{-6}{*}{\rotatebox[origin=c]{90}{\textbf{Total}}}
   & \textbf{\# Overlapping Spans}      & 69        & 54        & 178       & 301     \\
    \bottomrule
    \toprule
    \rowcolor{Gray}
   & \textbf{\# Posts}                  & \multicolumn{4}{r}{126,769} \\
     \rowcolor{Gray}
   & \textbf{\# Sentences}              & \multicolumn{4}{r}{3,195,585} \\
      \rowcolor{Gray}
   \multirow{-3}{*}{\rotatebox[origin=c]{90}{\textbf{\textsc{$\mathcal{U}$}}}}
   & \textbf{\# Tokens}                 & \multicolumn{4}{r}{460,484,670} \\
\bottomrule
    \end{tabular}
    }
    \caption{\textbf{Statistics of Dataset.} Indicated is the number of JPs across splits \& source and their respective number of sentences, tokens, and spans. The total is reported in the cyan column and rows. We report the overall statistics of the unlabeled JPs ($\mathcal{U}$) in the gray rows.}
    \label{tab:num_post}
\end{table}

\begin{figure}[ht]
    \centering
    \includegraphics[width=\linewidth]{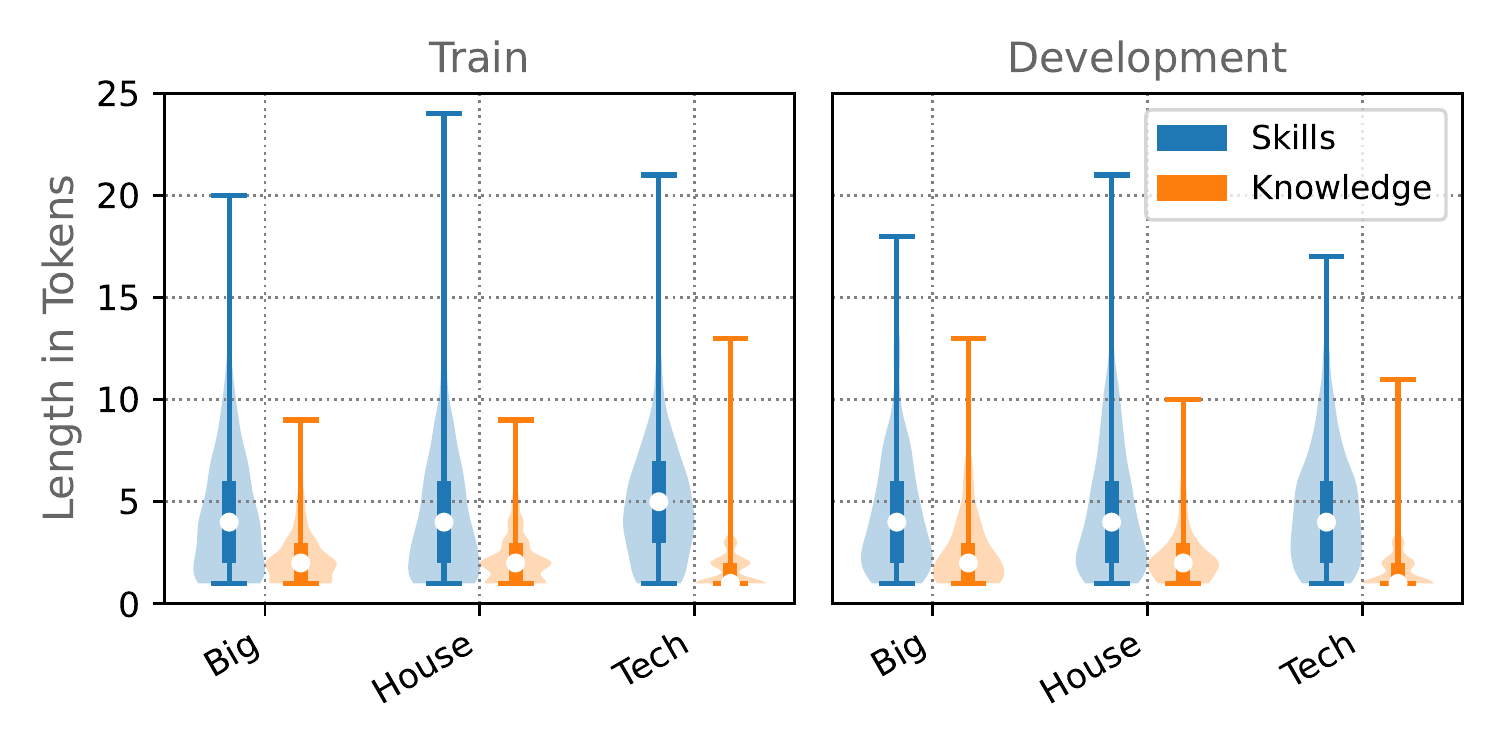}
    \caption{\textbf{Violin Plots of Annotated Components.} Indicated are the distributions regarding the length of spans in each type of annotated component (i.e., length of skills and knowledge components). The white dot is the median length, the bars range from the first quartile to the third quartile, and the colored line ranges from the lower adjacent value to the higher adjacent value.}
    \label{fig:violin}
\end{figure}

\paragraph{Data\footnote{Our data statement~\cite{bender-friedman-2018-data} can be found in~\cref{app:datastatement}.}} We continuously collected JPs via web data extraction between June 2020--September 2021.  Our JPs come from the three sources:

\begin{enumerate}
\itemsep0em
    \item \textbf{\bjo{}}: A large job platform with various types of JPs, with several type of positions;
    \item \textbf{\hou{}}: A \emph{static} in-house dataset consisting of similar types of jobs as \bjo{}. Dates range from 2012--2020;
    \item \textbf{\tech{}}: The StackOverflow JP platform that consisted mostly of technical jobs (e.g., developer positions).
\end{enumerate}

We release the anonymized raw data and annotations of the parts with permissible licenses, i.e., \hou{} (from a govermental agency which is our collaborator) and \tech{}.\footnote{Links to our data can be found at \url{https://github.com/kris927b/SkillSpan}.} For anonymization, we perform it via manual annotation of job-related sensitive and personal data regarding \texttt{Organization}, \texttt{Location}, \texttt{Contact}, and \texttt{Name} following the work by~\citet{jensen-etal-2021-de}.

\noindent
\cref{tab:num_post} shows the statistics of \textsc{SkillSpan}, with 391 annotated JPs from the three sources containing 14.5K sentences and 232.2K tokens. The unlabeled JPs (only to be released as pre-trained model) consist of 126.8K posts, 3.2M sentences, and 460.5M tokens. What stands out is that there are 2--5 times as many annotated knowledge components in \tech{} in contrast to the other sources, despite a similar amount of JPs. We expect this to be due the numerous KCs depicted in this domain (e.g., programming languages), while we observe considerably fewer soft skills (e.g., ``work flexibly''). The amount of skills is more balanced across the three sources. Furthermore, overlapping spans follow a consistent trend among splits, with the train split containing the most.

\paragraph{Data Annotation}

We annotate competences related to SKCs in two levels as illustrated in~\cref{fig:doccano}. We started the process in March 2021, with initial annotation rounds to construct and refine the annotation guidelines (as outlined further below). The annotation process spanned eight months in total.
Our final annotation guidelines can be found in~\cref{sec:ann_guide}. The guidelines were developed by largely following example spans given in the ESCO taxonomy. 
However, at this stage, we focus on span identification, and  we do not take the fine-grained taxonomy codes from ESCO for labeling the spans, 
leaving the mapping to ESCO and taxonomy enrichment as future work.

\paragraph{Further Details on the Annotation Process} The development of the annotation guidelines and our annotation process is depicted as follows: \circled{\textbf{1}} We wrote base guidelines derived from a small number of JPs. \circled{\textbf{2}} We had three pre-rounds consisting of three JPs each. After each round, we modified, improved and finalized the guidelines. \circled{\textbf{3}} Then, we had three longer-lasting annotation rounds consisting of 30 JPs each. We re-annotated the previous 11 JPs in \circled{1} and \circled{2}.
\circled{\textbf{4}} After these rounds, one of the annotators (the hired linguist) annotated JPs in batches of 50. The data in \circled{1}, \circled{2}, and \circled{3} was annotated by three annotators (101 JPs).

We used an open source text annotation tool named \textsc{Doccano} \cite{doccano}. 
There are around 57.5K tokens (approximately 4.6K sentences, in 101 job posts) that we calculated agreement on. 
The annotations were compared using Cohen's $\kappa$~\cite{fleiss1973equivalence} between pairs of annotators, and Fleiss' $\kappa$~\cite{fleiss1971measuring}, which generalises Cohen's $\kappa$ to more than two concurrent annotations. We consider two levels of $\kappa$ calculations: \textbf{\textsc{Token}} is calculated on the token level, comparing the agreement of annotators on each token (including non-entities) in the annotated dataset.~\textbf{\textsc{Span}} refers to the agreement between annotators on the exact span match over the surface string, regardless of the type of SKC, i.e., we only check the position of tag without regarding the type of the entity. 
The observed agreements scores over the three annotators from step \circled{3} are between 0.70--0.75 Fleiss' $\kappa$ for both levels of calculation which is considered a \emph{substantial agreement}~\cite{landis1977measurement} and a $\kappa$ value greater than 0.81 indicates \emph{almost perfect agreement}. Given the difficulty of this task, we consider the aforementioned $\kappa$ score to be strong. Particularly, we observed a large improvement in annotation agreement from the earlier rounds (step \circled{1} and \circled{2}), where our Fleiss' $\kappa$ was 0.59 on token-level and 0.62 for the span-level. 

Overall, we observe higher annotator agreement for knowledge components (3--5\% higher) compared to skills 
which tend to be longer. The \textsc{Tech} domain is the most consistent for agreement while \textsc{Big} shows more variation over rounds, likely due to the broader nature of the domains of JPs. 

\paragraph{Annotation Span Statistics}
A challenge of annotating spans is the length (i.e., boundary), SKCs being in different domains (e.g., business versus technical components), and frequently written differently, e.g., ``being able to work together'' v.s.\ ``teamwork''). \cref{fig:violin} shows the statistics of our annotations in violin plots. For the training set, the median length (white dot) of skills is around 4 for \bjo{} and \hou{}, for \tech{} this is a median of 5. In the development set, the median stays at length 4 across all sources. Another notable statistic is the upper and lower percentile of the length of skills and knowledge, indicated with the thick bars. Here, we highlight the fact that skill components could consist of many tokens, for example, up to length 7 in the \hou{} source split (see blue-colored violins). For knowledge components, the spans are usually shorter, where it is consistently below 5 tokens (see orange-colored violins). All statistics follow a similar distribution across train, development, and sources in terms of length and distribution. This gives a further strong indication that consistent annotation length has been conducted across splits and sources.

\begin{table}
\centering
    \resizebox{\linewidth}{!}{
    \begin{tabular}{l|lll}
    \toprule
& \textbf{\bjo{}} & \textbf{\hou{}} & \textbf{\tech{}}\\
\midrule
\multirow{5}{*}{\rotatebox[origin=l]{90}{\textbf{\textsc{Skill}}}}
    & ambitious              & structured           & hands-on                      \\
    & proactive              & teaching             & communication skills             \\
    & work independently     & communication skills & leadership                    \\
    & attention to detail    & project management   & passionate                    \\
    & motivated              & drive                & open-minded                   \\
    \midrule
\multirow{5}{*}{\rotatebox[origin=l]{90}{\textbf{\textsc{knowledge}}}}
    & full uk driving licence   & english              & java        \\
    & sap energy assessments    & supply chain         & javascript  \\
    & right to work in the uk   & project management   & aws         \\
    & sen                       & powders              & docker      \\
    & acca/aca                  & machine learning     & node.js     \\
    \bottomrule
    \end{tabular}}
    \caption{\textbf{Most Frequent Skills in the Development Data.} Top-5 skill components in our data in terms of frequency on different sources. A larger example can be found in~\cref{tab:freq-skill} and~\cref{tab:freq-knowledge} (\cref{app:quali}).}
    \label{tab:freq-top5}
\end{table}

\paragraph{Qualitative Analysis of Annotations} 
Qualitative differences in SKCs over the three sources are shown (lowercased) in \cref{tab:freq-top5}. With respect to skill components, all sources follow a similar usage of skills. The annotated skills mostly relate to the \emph{attitude} of a person 
and hence mostly consist of soft skills.
With respect to knowledge components, we observe differences between the three sources. First, on the source-level, the knowledge components vastly differ between \bjo{} and \tech{}. \bjo{} postings seem to cover more \emph{business} related components, whereas \tech{} has more \emph{engineering} components. \hou{} seems to be a mix of the other two sources. Lastly, note that both the skill and knowledge components between the splits diverge in terms of the type of annotated spans, which indicates a variation in the annotated components. We show the top--10 skills annotated in the train, development, and test splits for SKCs in~\cref{app:quali}.
From a syntactic perspective, skills frequently consist of noun phrases, verb phrases, or adjectives (for soft skills). Knowledge components usually consists of nouns or proper nouns, such as ``python'', ``java'', and so forth. 

\begin{figure*}
    \centering
    \includegraphics[width=.9\linewidth]{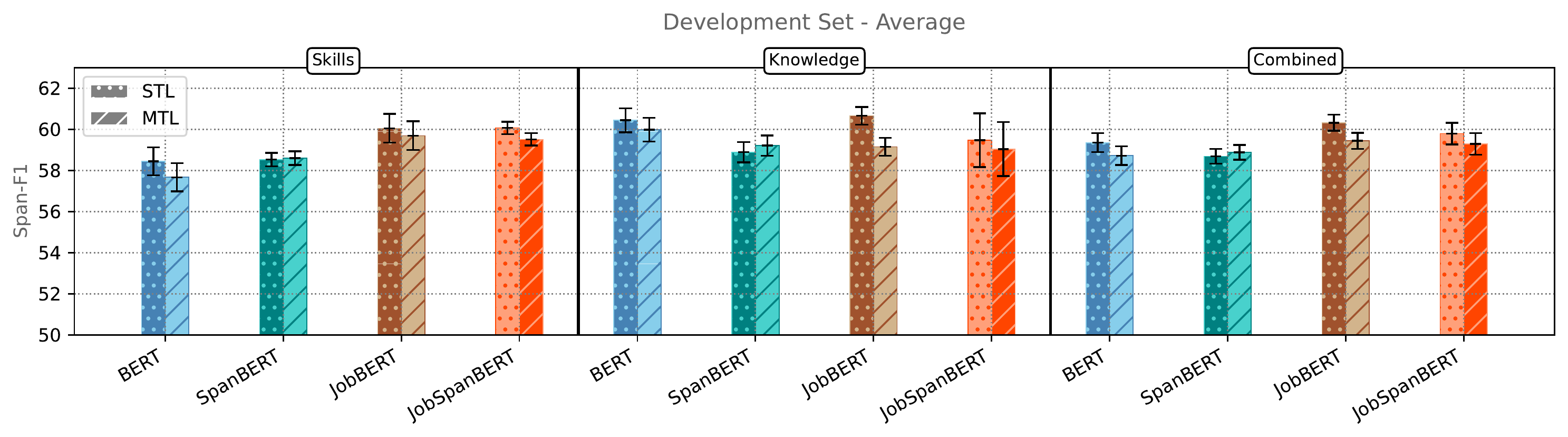}
    \includegraphics[width=.9\linewidth]{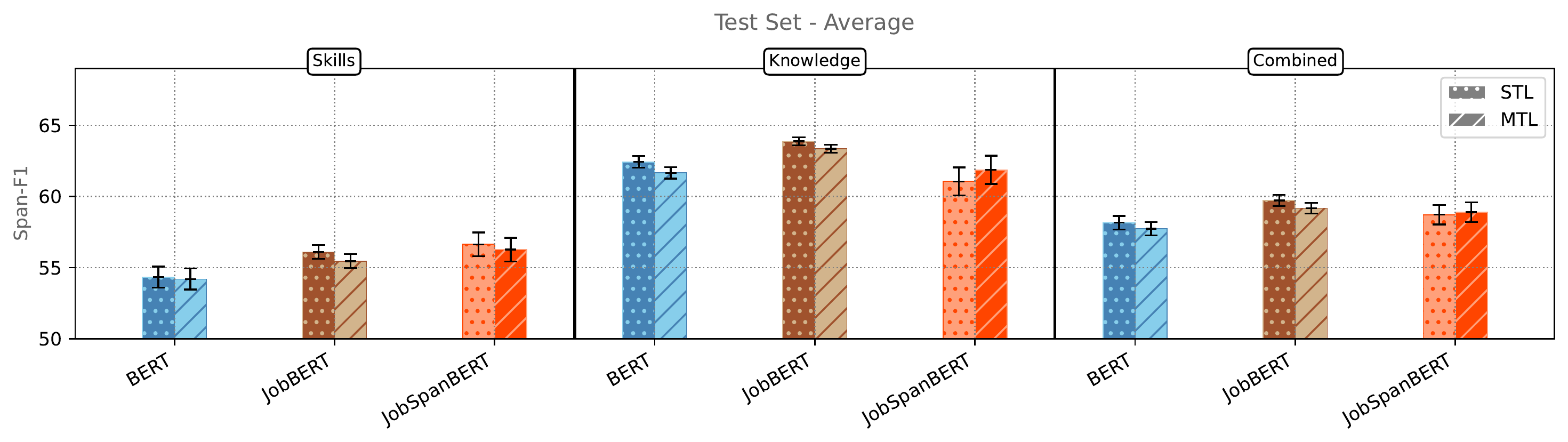}
 
    \caption{\textbf{Performance of Models.} We test the models on \textbf{\textsc{Skills}}, \textbf{\textsc{Knowledge}}, and \textbf{\textsc{Combined}}. We report the span-F1 and standard deviation (error bars) of runs on five random seeds. \textbf{Note that the y-axis starts from 50 span-F1}. \textbf{STL} indicates single-task learning and \textbf{MTL} indicates the multi-task model. Differences can be seen on the test set: JobSpanBERT performs best on \textsc{Skills}, JobBERT is best on \textsc{Knowledge}, and JobBERT achieves best in \textsc{Combined}. Exact numbers of the plots are in~\cref{tab:results} (\cref{app:source-perf}).}
    \label{fig:results}
\end{figure*}

\section{Experimental Setup}

The task of SE is formulated as a sequence labeling problem. Formally, we consider a set of JPs $\mathcal{D}$, where $d \in \mathcal{D}$ is a set of sequences (i.e., entire JPs) with the $i^\text{th}$ input sequence $\mathcal{X}^i_{d} = \{x_1, x_2, ..., x_T\}$ and a target sequence of \texttt{BIO}-labels $\mathcal{Y}^i_{d} = \{y_1, y_2, ..., y_T\}$ (e.g., ``\texttt{B-SKILL}'', ``\texttt{I-KNOWLEDGE}'', ``\texttt{O}''). The goal is to use $\mathcal{D}$ to train a sequence labeling algorithm $h: \mathcal{X} \mapsto \mathcal{Y}$ to accurately predict entity spans by assigning an output label $y_t$ to each token $x_t$. 

As baseline we consider BERT and we investigate more recent variants, and we also train models from scratch. Models are chosen due to their state-of-the-art performance, or in particular, for their strong performance on  longer spans.

\paragraph{\bertb~\cite{devlin2019bert}} An out-of-the-box \bertb{} model (\texttt{bert-base-cased}) from the \texttt{HuggingFace} library~\cite{wolf-etal-2020-transformers} functioning as a baseline.

\paragraph{SpanBERT~\cite{joshi2020spanbert}} A BERT-style model that focuses on span representations as opposed to single token representations. SpanBERT is trained by masking contiguous spans of tokens and optimizing two objectives: (1) masked language modeling, which predicts each masked token from its own vector representation. (2) The span boundary objective, which predicts each masked token from the representations of the unmasked tokens at the start and end of the masked span.

We train a SpanBERT\textsubscript{base} model from scratch on the BooksCorpus~\cite{Zhu_2015_ICCV} and English Wikipedia using cased Wordpiece tokens~\cite{wu2016google}. We use AdamW~\cite{kingma2014adam} for 2.4M training steps with batches of 256 sequences of length 512. The learning rate is warmed up for 10K steps to a maximum value of 1$e$-4, after which it has a decoupled weight decay~\cite{loshchilov2017decoupled} of 0.1. We add a dropout rate of 0.1 across all layers. Pretraining was done on a v3-8 TPU on the GCP and took 14 days to complete. We take the official TensorFlow implementation of SpanBERT by~\citet{ram2021few}.

\paragraph{JobBERT\footnote{\url{https://huggingface.co/jjzha/jobbert-base-cased}}} We apply domain-adaptive pre-training~\cite{gururangan2020don} to a \bertb~model using the 3.2M unlabeled JP sentences~(\cref{tab:num_post}). Domain-adaptive pre-training relates to the continued self-supervised pre-training of a large language model on domain-specific text. This approach improves the modeling of text for downstream tasks within the domain. We continue training the BERT model for three epochs (default in \texttt{HuggingFace}) with a batch size of 16.

\paragraph{JobSpanBERT\footnote{\url{https://huggingface.co/jjzha/jobspanbert-base-cased}}} We apply domain-adaptive pre-training to our SpanBERT on 3.2M unlabeled JP sentences. We keep parameters identical to the vanilla SpanBERT, but change the number of steps to 40K to have three passes over the unlabeled data.

\begin{figure*}
    \centering
    \includegraphics[width=.32\linewidth]{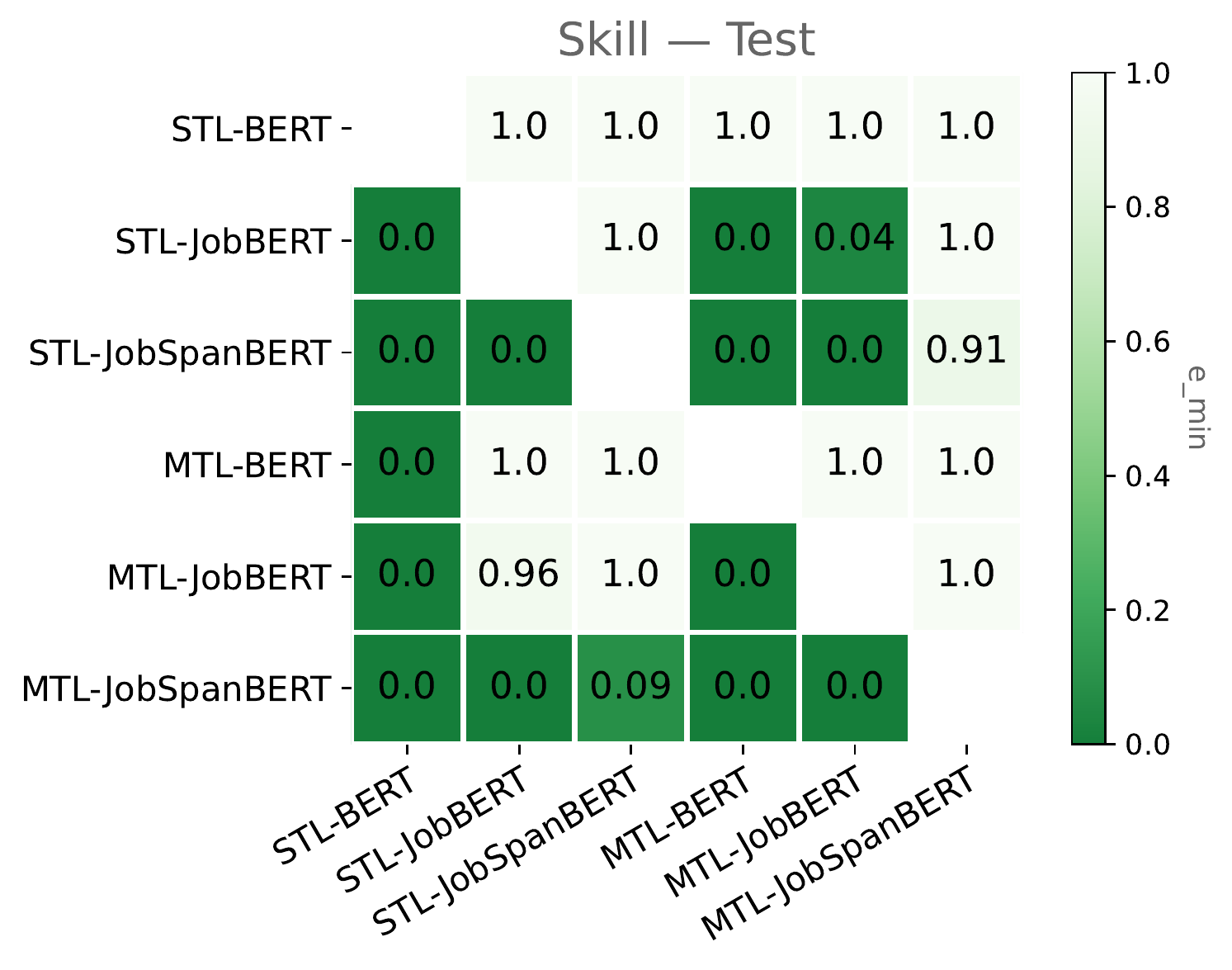}
    \includegraphics[width=.32\linewidth]{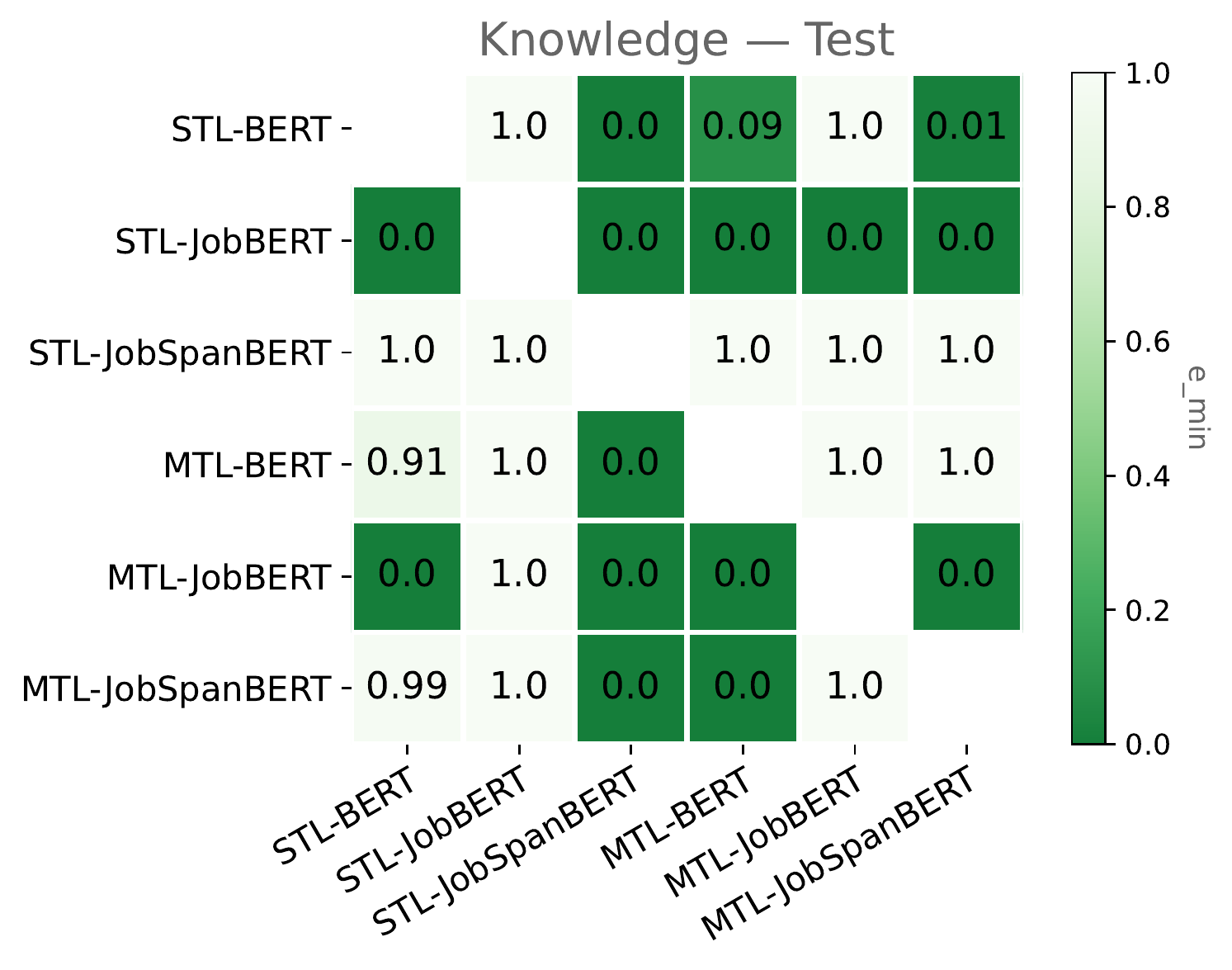}
    \includegraphics[width=.32\linewidth]{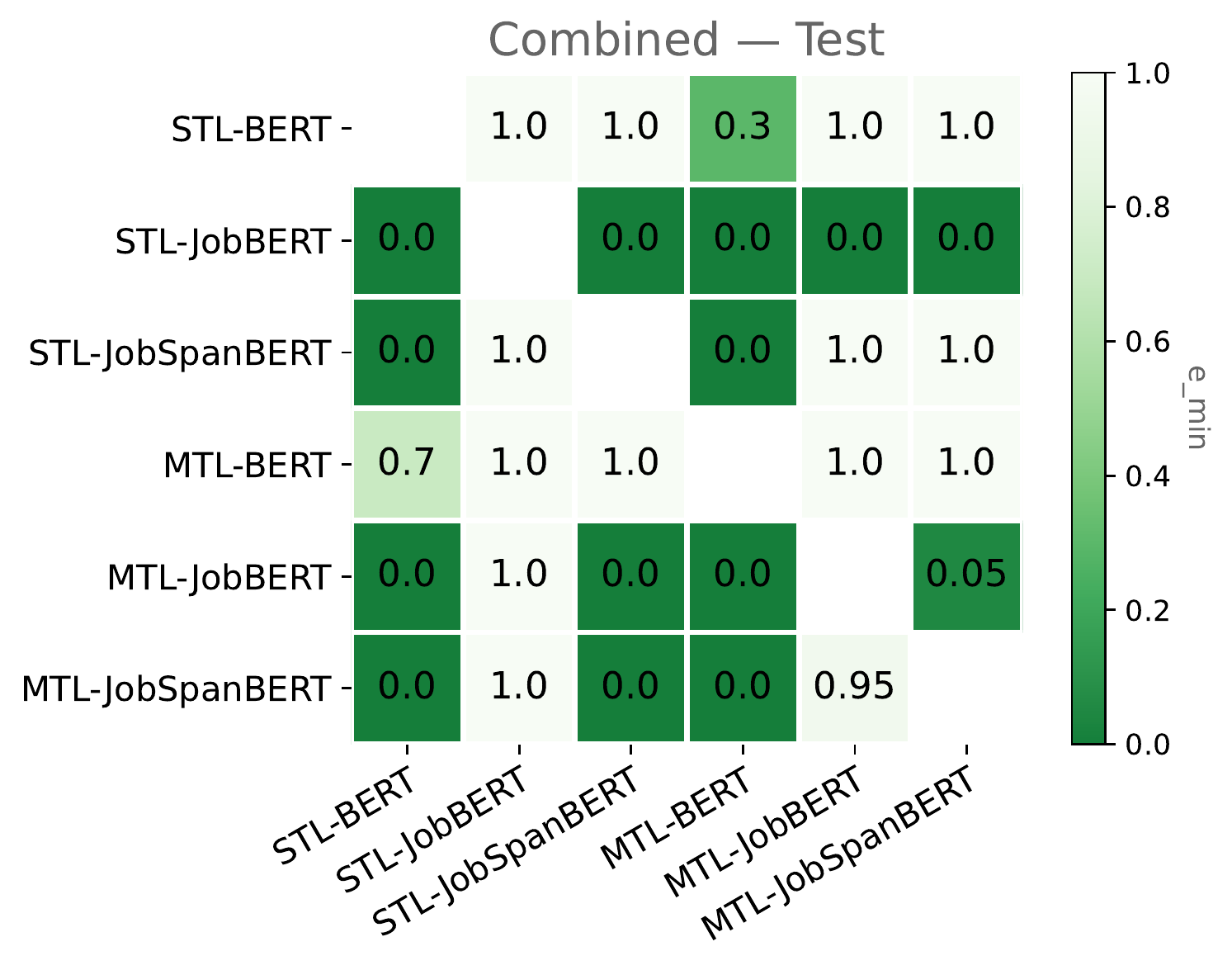}
    \caption{\textbf{Almost Stochastic Order Scores of the Test Set.}
    ASO scores expressed in $\epsilon_\text{min}$.
    The significance level $\alpha =$ 0.05 is adjusted accordingly by using the Bonferroni correction~\citep{bonferroni1936teoria}. Read from row to column: E.g., in \textsc{Combined} STL-JobBERT (row) is stochastically dominant over STL-\bertb{} (column) with $\epsilon_\text{min}$ of 0.00.}
    \label{fig:significance-test}
\end{figure*}

\paragraph{Experiments}
We have 391 annotated JPs (\cref{tab:num_post}) that we divide across three splits: Train, dev.\, and test set. We use 101 JPs that all three annotators annotated as the gold standard test set with aggregated annotations via majority voting. The 101 postings are divided between the sources as: 36 \bjo{}, 33 \hou{}, and 32 \tech{}. The remaining 290 JPs were annotated by one annotator. We use 90 JPs (30 from each source, namely \bjo{}, \hou{}, and \tech{}) as the dev.\ set. The remaining 200 JPs are used as the train set. The sources in the train set are divided into 60 \bjo{}, 60 \hou{}, and 80 \tech{}.

\paragraph{Setup} The data is structured as \texttt{CoNLL} format~\cite{tjong2003introduction}. For the nested annotations, the skill tags are appearing only in the first column and the knowledge tags are only appearing in the second column of the file and they are allowed to overlap with each other. 
We perform experiments with single-task learning (STL) on either the skill or knowledge components, MTL for predicting both skill and knowledge tags at the same time, while evaluating the MTL models also on either skills or knowledge components. We used a single joint MTL model with hard-parameter sharing~\cite{caruana1997multitask}.
All models are with a final Conditional Random Field~\citep[CRF;\ ][]{lafferty2001conditional} layer. Earlier research, such as~\citet{souza2019portuguese, jensen-etal-2021-de} show that BERT models with a CRF-layer improve or perform similarly to its simpler variants when comparing the overall F1 and make no tagging errors (e.g., \texttt{B}-tag follows \texttt{I}-tag). In the case of MTL we use one for each tag type (skill and knowledge). In the STL experiments we use one CRF for the given tag type.

We use the \textsc{MaChAmp} toolkit~\cite{van-der-goot-etal-2021-massive} for our experiments. For each setup we do five runs (i.e., five random seeds).\footnote{For reproducibility, we refer to~\cref{app:hyper}.} 
For evaluation we use span-level precision, recall, and F1, where the F1 for the MTL setting is calculated as described in~\citet{benikova2014nosta}.

\section{Results}

The results of the experiments are given in \cref{fig:results}. We show the average performance of each model in F1 and respective standard deviation over the development and test split.
Exact scores on each source split and other metric details are provided in~\cref{app:source-perf}. As mentioned before, we experiment with the following settings: \textbf{\textsc{Skill}}, we train and predict only on skills. \textbf{\textsc{Knowledge}}, train and only predict for knowledge. \textbf{\textsc{Combined}}, we merge the STL predictions of both skills and knowledge. We also train the models in an MTL setting, predicting both skills and knowledge simultaneously. We evaluate the MTL model on both \textbf{\textsc{Skill}} and \textbf{\textsc{Knowledge}} separately, and also compare it against the aggregated STL predictions.

\paragraph{Performance on Development Set} In \cref{fig:results}, we show the results on the development set in the upper plot. We observe similar performance between the domain-adapted STL models---JobBERT and JobSpanBERT---have similar span-F1 for \textsc{Skill}: \std{60.05}{0.70} vs.\ \std{60.07}{0.70}. In contrast, for \textsc{Knowledge}, \bertb{} and JobBERT are closest in predictive performance: \std{60.44}{0.58} vs.\ \std{60.66}{0.43}. In the \textsc{Combined} setting, JobBERT performs highest with a span-F1 of \std{60.32}{0.39}. On average, JobBERT performs best over all three settings.
Surprisingly, the models for both \textsc{Skill} and \textsc{Knowledge} perform similarly (around 60 span-F1), despite the sources' differences in properties and length~\cref{fig:violin}.
In addition, we find that MTL is not performing better than STL across sources. 
For exact numbers and source-level (i.e., \bjo{}, \hou{}, \tech{}), we refer to~\cref{app:source-perf}.

\paragraph{Performance on Test Set} We select the best performing models in the development set evaluation and apply it to the test set. Results are in~\cref{fig:results} in the bottom plot. Since JobBERT and JobSpanBERT are performing similarly, we apply both to the test set and \bertb{}. 
We observe a deviation from the development set to the test set: JobSpanBERT \std{60.07}{0.30}$\rightarrow$\std{56.64}{0.83} on \textsc{Skill}, JobBERT \std{60.66}{0.43}$\rightarrow$\std{63.88}{0.28}
on \textsc{Knowledge}. For \textsc{Combined}, JobBERT performs slightly worse: \std{60.32}{0.39}$\rightarrow$\std{59.73}{0.38}. Similar to the development set, we find that on all three methods of evaluation (i.e., \textsc{Skill}, \textsc{Knowledge}, and \textsc{Combined}), STL still outperforms MTL. For \textsc{Skill} and \textsc{Knowledge}, STL is almost stochastically dominant over MTL (i.e., significant), and for \textsc{Combined} there is stochastic dominance of STL over MTL, indicated in the next paragraph. 

\paragraph{Significance}
We compare all pairs of models based on five random seeds each using Almost Stochastic Order~\citep[ASO;\  ][]{dror2019deep} tests with a confidence level of $\alpha =$ 0.05.
The ASO scores of the test set are indicated in~\cref{fig:significance-test}. 
We show that MTL-JobSpanBERT for \textsc{Skill} shows almost stochastic dominance ($\epsilon_\text{min} < $ 0.5) over all other models. For \textsc{Knowledge} and \textsc{Combined}, We show that STL-JobBERT is stochastically dominant ($\epsilon_\text{min} = $ 0.0) over \emph{all} the other models.
For more details, we refer to~\cref{app:sign-per-source} for ASO scores on the development set.

\section{Discussion}

\paragraph{What Did Not Work} Additionally, we experiment whether representing the entire JP for extracting tokens yields better results than the experiments so far, which were sentence-by-sentence processing setups. To handle entire JPs and hence much longer sequences we use a pre-trained Longformer$_\text{base}$~\cite{beltagy2020longformer} model. The document length we use in the experiments is 4096 tokens. Results of the Longformer on the test set are lower: For skills, JobSpanBERT against Longformer results in \std{56.64}{0.83} vs.\ \std{52.55}{2.39}. For \textsc{Knowledge}, JobBERT against Longformer shows \std{63.88}{0.28} vs.\ \std{57.26}{1.05}. Last, for \textsc{Combined}, JobBERT against Longformer results in \std{59.73}{0.38} vs.\ \std{55.05}{0.71}. This drop in performance is difficult to attribute to a concrete reason: e.g., the Longformer is trained on more varied sources than BERT, but not specifically for JPs, which may have contributed to this gap.
Since the vanilla Longformer already performs worse than \bertb~overall, we did not opt to apply domain-adaptive pre-training. Overall, we show that representing the full JP is not beneficial for SE, at least not in the Longformer setup tested here.

\paragraph{Continuous Pretraining helps SE}

As previously mentioned, due to the domain specialization of the domain-adapted pre-trained BERT models, they predict more skills and frequently perform better in terms of precision, recall, and F1 as compared to their non-adaptive counterparts. This is especially encouraging as we confirm findings that continuous pre-training helps to adapt models to a specific domain~\cite{alsentzer-etal-2019-publicly, lee2020biobert, gururangan2020don, nguyen-etal-2020-bertweet}. However, there are exceptions. Particularly in~\cref{tab:results} on \textsc{Test} for \textsc{Knowledge}, \bertb{} comes closer in predictive performance to JobBERT (difference of 1.5 F1) than on \textsc{Skills}. Our intuition is that knowledge components are often already in the pre-training data (e.g., Wikipedia pages of certain competences like Python, Java etc.) and therefore adaptive pre-training does not substantially boost performance.

\begin{figure}
    \centering
    \includegraphics[width=\linewidth]{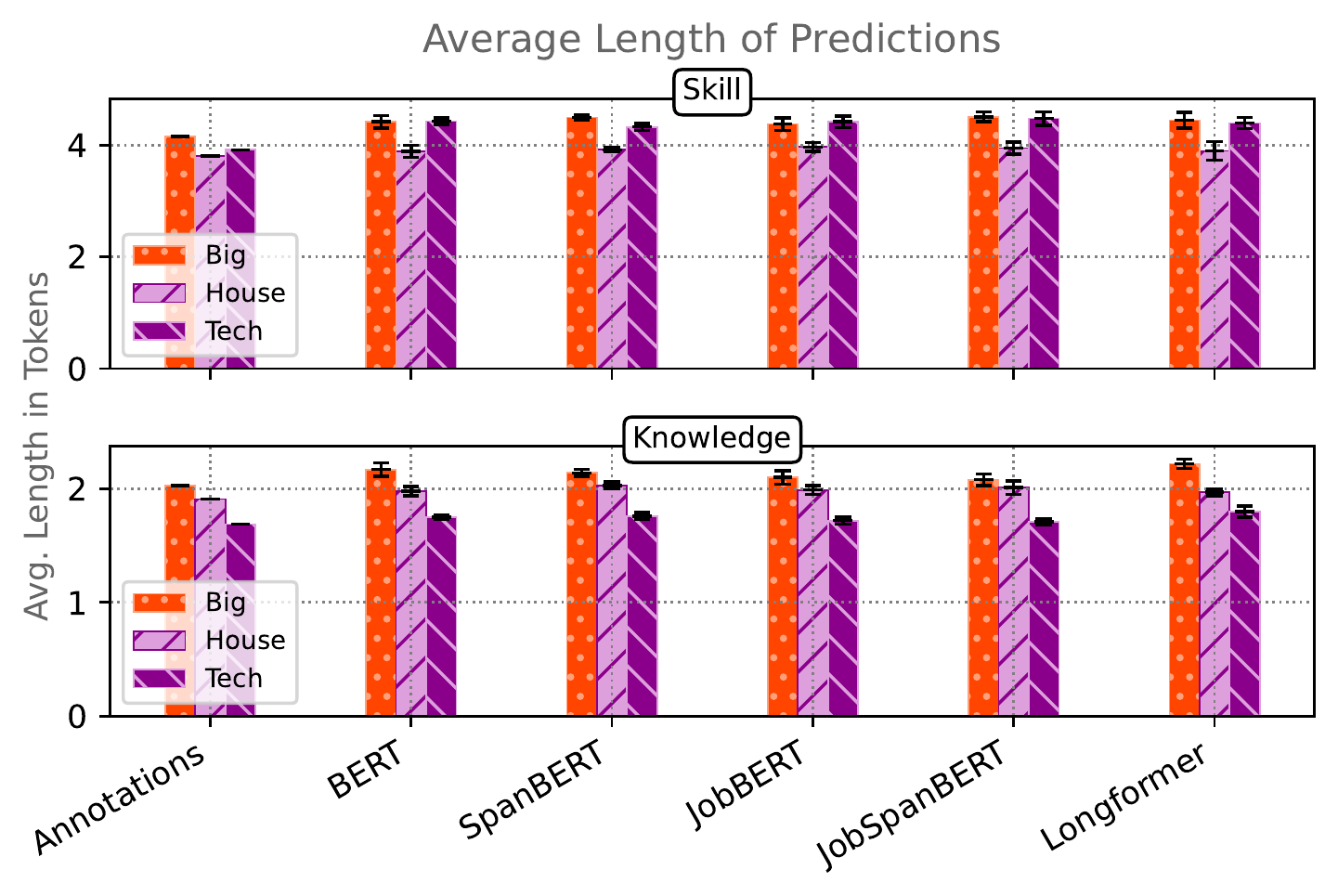}
    \caption{\textbf{Average Length of Predictions of Single Models.} We show the average length of the predictions versus the length of our annotated skills and knowledge components on the \emph{test set} and the total number of predicted skills and knowledge tags in each respective split (\#). There is a consistent trend over the three sources.}
    \label{fig:avglength}
\end{figure}

\paragraph{Difference in Length of Predictions} 
The main motivation of selecting models optimized for long spans was the length of the annotations (\cref{fig:violin}). We investigate the average length of predictions of each model~(\cref{fig:avglength}) to find out whether the models that are adapted to handle longer sequences truly predict longer spans. Interestingly, the average length of predicted skills are longer than the annotations over all three sources. There is a consistent trend among \textsc{Skill}: \bjo{} and \tech{} have similar length over predictions ($>$4), while \hou{} is usually lower than length 3.
For both \bjo{} and \tech{}, JobSpanBERT predicts the longest skill spans (4.51 and 4.48 respectively). We suspect due to the domain-adaptive pre-training on JPs, it improved the span prediction performance. In contrast, the Longformer predicts shorter spans. Note that the Longformer is not domain-adapted to JPs. 

Regarding \textsc{Knowledge}, there is also a consistent trend: \bjo{} has the overall longest prediction length while \tech{} has the lowest. The Longformer predicts the longest spans on average for \bjo{} and \tech{}. Knowledge components are representative of a normal-length NER task and might not need a specialized model for long sequences. We show the exact numbers in~\cref{tab:avglength} (\cref{app:source-perf}) and the number of predicted \textsc{Skill} and \textsc{Knowledge}: JobBERT and JobSpanBERT have higher recall than the other models.

\begin{figure*}[t]
    \centering
    \includegraphics[width=.85\linewidth]{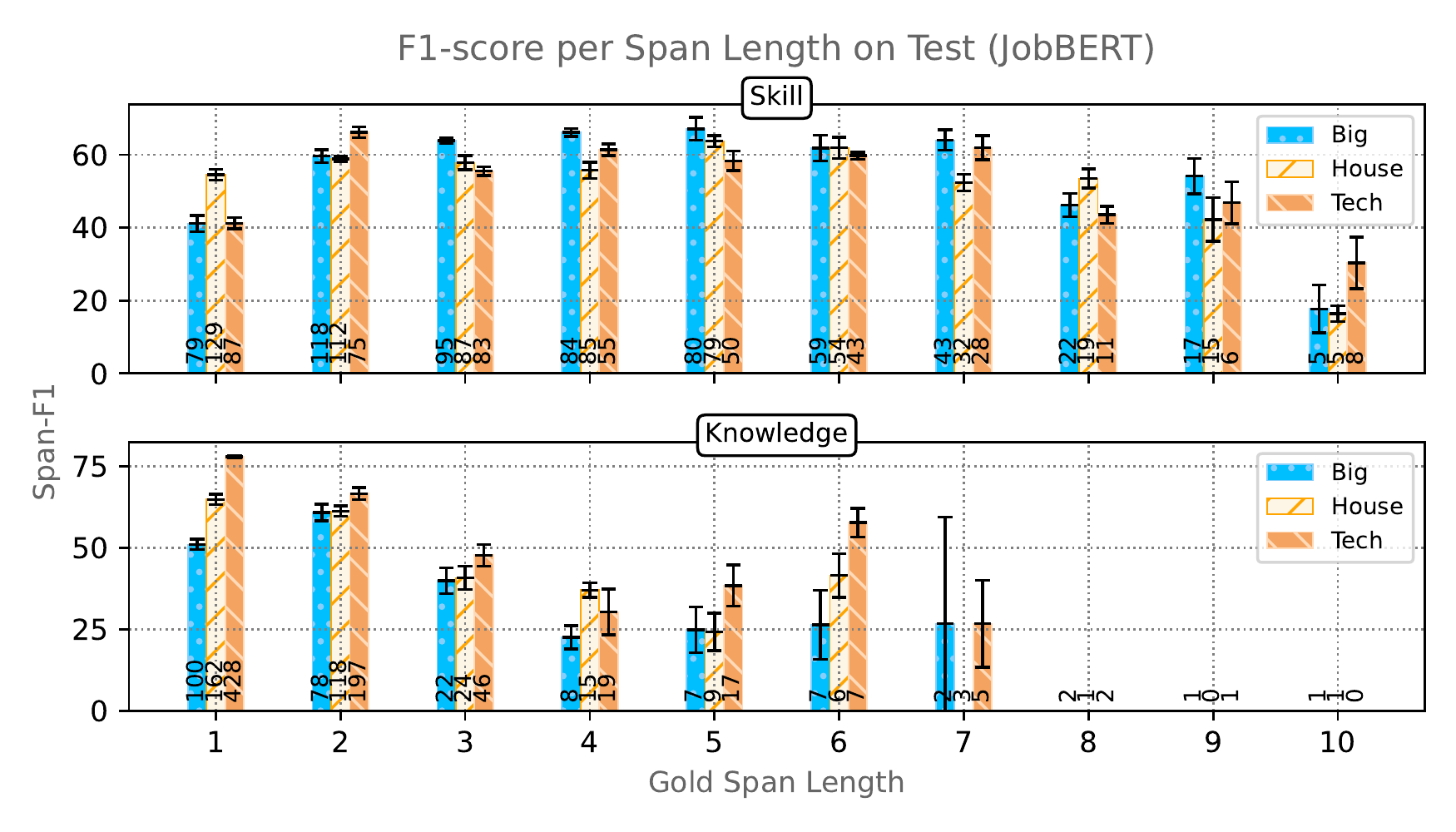}
    \caption{\textbf{Average Span-F1 per Span Length.} We bucket the performance of JobBERT according to the length of the spans until 10 tokens and show the performance on each length, averaged over five random seeds. Indicated per bar is the support. The model performs best on medium-length skill spans (i.e., spans with token length of 4-5). For knowledge spans, on average, it performs best on short-length spans (i.e., spans with token length of 1-2).}
    \label{fig:predlength}
\end{figure*}

\paragraph{Performance per Span Length}
\textsc{Skills} are generally longer than \textsc{Knowledge} components in our dataset (\cref{fig:violin}). The previous overall results on the test set (\cref{fig:results}) show that performance on \textsc{Skill} is substantially lower than \textsc{Knowledge}. We therefore investigate whether this performance difference is attributed to the longer spans in \textsc{Skill}. In~\cref{fig:predlength}, we show the average performance of the best performing model (JobBERT) on the three sources (test set) based on the gold span length, until a length of 10.

In \textsc{Skill} components (upper plot), we see much support for spans with length 1 and 2, which then lowers once the spans become longer. Spans with length of 1 shows low performance on \bjo{} and \tech{} (around 40 span-F1), which influences the total span-F1. Short skills are usually soft skills, such as ``passionate'', which can be used as a skill or not. This might confuse the model. In contrast, performance effectively stays similar (around 60 span-F1) for span length of 2 till 7 for all sources. Afterwards, it drops in performance. Thus, the weak performance on \textsc{Skill} seem to be due to lower performance on the short spans.

For the \textsc{Knowledge} components (lower plot), they are generally shorter.  We see that there is a gap in support between the sources, \tech{} has a larger number of gold labels compared to \bjo{} and \hou{}. Unlike soft skills, KCs usually consist of proper nouns such as ``Python'', ``Java'', and so forth, which connects to the high performance on \tech{} (around 76 span-F1). Furthermore, support for spans longer than 2 drops considerably. In this case, if the model predicts a couple of instances correctly, it would substantially increase span-F1. Contrary to \textsc{Skill}, high performance of \textsc{Knowledge} can be attributed to its strong performance on short spans.

\section{Conclusion}
We present a novel dataset for skill extraction on English job postings--- \textsc{SkillSpan}---and domain-adapted BERT models---JobBERT and JobSpanBERT. We outline the dataset and annotation guidelines, created for hard \emph{and} soft skills annotation on the \emph{span-level}.
Our analysis shows that domain-adaptive pre-training helps to improve performance on the task for both skills and knowledge components. Our domain-adapted JobSpanBERT performs best on skills and JobBERT on knowledge. Both models achieve almost stochastic dominance over all other models for skills and knowledge extraction, whereas JobBERT in the STL setting achieves stochastic dominance over other models. 

With the rapid emergence of new competences, our new approach of skill extraction has future potential, e.g., to enrich knowledge bases such as ESCO with unseen skills or knowledge components, and in general, contribute to providing insights into labor market dynamics. We hope our dataset encourages research into this emerging area of computational job market analysis. 

\section*{Acknowledgements}
We thank Google’s TFRC for their support in providing TPUs for this research. Furthermore, we thank the NLPnorth group for feedback on an earlier version of this paper---in particular, Elisa Bassignana and Max M\"uller-Eberstein for insightful discussions. We would also like to thank the anonymous reviewers for their comments to improve this paper. Last, we also thank NVIDIA and the ITU High-performance Computing cluster for computing resources. This research is supported by the Independent Research Fund Denmark (DFF) grant 9131-00019B.

\bibliography{anthology,custom}
\bibliographystyle{acl_natbib}

\appendix
\vfill
\section{Data Statement \textsc{SkillSpan}}\label{app:datastatement}
Following~\citet{bender-friedman-2018-data}, the following outlines the data statement for \textsc{SkillSpan}:
\begin{enumerate}[A.]
    \item \textsc{Curation Rationale}: Collection of job postings in the English language for span-level sequence labeling, to study the impact of sequence labeling on the extraction of skill and knowledge components from job postings.
    \item \textsc{Language Variety}: The non-canonical data was collected from the StackOverflow job posting platform, an in-house job posting collection from our national labor agency collaboration partner (\emph{which will be elaborated upon acceptance}), and web extracted job postings from a large job posting platform. US (en-US) and British (en-GB) English are involved.
    \item \textsc{Speaker Demographic}: Gender, age, race-ethnicity, socioeconomic status are unknown.
    \item \textsc{Annotator Demographic}: Three hired project participants (age range: 25--30), gender: one female and two males, white European and Asian (non-Hispanic). Native language: Danish, Dutch. Socioeconomic status: higher-education students. Female annotator is a professional annotator with a background in Linguistics and the two males with a background in Computer Science.
    \item \textsc{Speech Situation}: Standard American or British English used in job postings. Time frame of the data is between 2012--2021.
    \item \textsc{Text Characteristics}: Sentences are from job postings posted on official job vacancy platforms.
    \item \textsc{Recording Quality}: N/A.
    \item \textsc{Other}: N/A.
    \item \textsc{Provenance Appendix}: \emph{More info will be released upon acceptance.}
\end{enumerate}

\clearpage

\section{Annotation Guidelines}\label{sec:ann_guide}

\subsection{Span Specifications}
\guide{
\textbf{Legend:} \colorbox{pink}{Skill}, \colorbox{yellow}{Knowledge}, ``•'' indicates an example sentence.
}

\guide{
    1. A skill starts with a \textbf{VERB}, otherwise (\textbf{ADJECTIVE}) + \textbf{NOUN}\\\\
    1.1 Modal verbs are not tagged:
    \begin{itemize}\setlength\itemsep{0em}
        \item \st{Can} \sk{put personal touch on the menu}.
        \item \st{Will} \sk{train new staff}.
    \end{itemize}
}

\guide{
2. Split up phrases with prepositions and/or conjunctions\\\\
2.1 \textbf{Unless} the conjunction coordinates two nouns functioning as one argument:
\begin{itemize}
    \item \sk{Coordinate parties and conferences}.
\end{itemize}
2.2 Do not tag skills with anaphoric pronouns, only tag preceding skill:
\begin{itemize}
    \item \sk{Prioritizing tasks} and identifying those that are most important.
\end{itemize}
2.3 Split nouns and adjectives that are coordinated if they do not have a verb attached:
\begin{itemize}
    \item Be \sk{inquisitive} and \sk{proactive}.
    \item Prior in-house experience with \kn{media}, \kn{publishing} or \kn{internet companies}.
\end{itemize}
2.4 If there is a listing of skill tags and they lead up to different subtasks, we split them:
\begin{itemize}
    \item \sk{keep up the high level of quality in our team} through \sk{reviews}, \sk{pairing} and \sk{mentoring}.
\end{itemize}
}

\guide{
    3. If there is relevant information appended after irrelevant information (e.g., info specific to a company) we try to make the skill as short as possible:
    \begin{itemize}
        \item \sk{providing the best solution} \st{for Siemens Gamesa in a very} \sk{structured} and \sk{analytic} manner.
    \end{itemize}
}

\guide{
    4. Note also the words skills and knowledge can be included in the span of the component if leaving it out makes it nonsensical:
    \begin{itemize}
        \item \sk{personal skills} → just [personal] would make it nonsensical.
    \end{itemize}
}

\guide{
    5. Parentheses after a skill tag are included if they elaborate the component before them or if they are an abbreviation of the component.
}
\clearpage
\guide{
6. \textbf{Inclusion of adverbials in components}. Adverbials are included if it concerns the manner of doing something. All others are excluded:
\begin{itemize}
    \item   \st{like to} \sk{solve technical challenges \underline{independently}}.
    \item   \sk{communicates \underline{openly}}.
    \item   \sk{striving for the best} \st{in all that they do}.
    \item   \sk{Deliver first class customer service} \st{to our guests}.
    \item   \sk{Making the right decisions} \st{early in the process}.
\end{itemize}
}   
\guide{
7. \textbf{Attitudes as skills.} We annotate attitudes as a skill:
\begin{itemize}
    \item \st{a} \sk{can-do-approach} → we leave out articles from the attitude.
\end{itemize}
8. Attitudes are not tagged if they contain skill/knowledge components---then only the span of the skill is tagged.

\begin{itemize}
    \item \st{like to} \sk{solve technical challenges independently}.
    \item \st{Passion for} \kn{automation}.
    \item \st{enjoy} \sk{working in a team}.
\end{itemize}
}

\guide{
9. \textbf{Miscellaneous:}\\\\
9.1 Do \underline{not} tag ironic skills (e.g., lazy).\\\\
9.2 Avoid nesting of skills, annotate it as one span.\\\\
9.3  We annotate all skills that are part of sections such as ``requirements'', ``good-to-haves'', ``great-to-knows'', ``optionals'', ``after this $x$ months of training you'll be able to...'', ``At the job you're going to...''.\\\\
9.4 When there is a general standard that can be added to the skill, we add these: 
\begin{itemize}
    \item \sk{Process payments according to the [...] standards}.
\end{itemize}
}
\clearpage

\subsection{Knowledge Specifications}\label{sub:knowledge}
\guide{
1. \textbf{Rule-of-thumb}: knowledge is something that one possesses, and cannot (usually) physically execute:
\begin{itemize}
    \item \kn{Python} (programming language).
    \item \kn{Business}.
    \item \kn{Relational Databases}.
\end{itemize}
}

\guide{
2. If there is a component between parentheses that belongs to the knowledge component, we add it:
\begin{itemize}
    \item \kn{(non-) relational databases}.
    \item \kn{Driver License (UK/EU)}.
\end{itemize}
}

\guide{
3. \textbf{Licenses and certifications}: We add the additional words ``certificate'', ``card'', ``license'', et cetera. to the knowledge component.
}

\guide{
4. If the knowledge component looks like a skill, but the preceding verb is vague and empty (e.g., \textit{follow}, \textit{use}, \textit{comply with}, \textit{work with}) → only tag the knowledge component:
\begin{itemize}
    \item \st{Comply with} \kn{Food Code of Practice}.
    \item \st{Work with} \kn{AWS infrastructure}.
\end{itemize}
}

\guide{
5. We annotate only specified knowledge components:
\begin{itemize}
    \item \kn{MongoDB} or other \kn{NoSQL database}.
    \item \kn{JEST} \st{or other test libraries}. → ``other test libraries'' is under-specified.
\end{itemize}
}

\guide{
6. Knowledge components can be nested in skill components.
\begin{itemize}
    \item \sk{Design, execution and analysis of \kn{phosphoproteomics} experiments}.
\end{itemize}
}

\guide{
7. If all components coordinate/share one knowledge tag, we annotate it as one:
\begin{itemize}
    \item \kn{application, data and infrastructure architecture}. → The knowledge tags coordinate to ``architecture''. 
    \item \kn{chemical/biochemical engineering}.
\end{itemize}
}

\guide{
8. If there is a listing of knowledge tags, we annotate all knowledge tags separately:
\begin{itemize}
    \item \kn{Bachelor Degree} in \kn{Mathematics}, \kn{Computer Science}, or \kn{Engineering}.
\end{itemize}
}

\clearpage

\subsection{Other Specifications}
\guide{
1. \textbf{Rule-of-thumb}: If in doubt, annotate it as a skill.
}

\guide{
2. We are preferring skills over knowledge components.
}

\guide{
3. We prioritize skills over attitudes; if there is a skill within the attitude, only tag the skill:
\begin{itemize}
    \item \st{Passionate around} \sk{solving business problems} through \\ \kn{innovation \& engineering practices}.
\end{itemize}

}

\guide{
4. Skill or knowledge components in the top headlines of the JP are not tagged (e.g., title of a JP). If it is a sub-headline or in the rest of the posting, tag it.
}

\guide{
5. We try to keep the skill/knowledge components as short as possible (i.e., exclude information at the end if it makes it too specific for the job).
}

\guide{
6. We do not include ``fluff'' and ``triggers'' (i.e., words that indicate a skill or knowledge component will follow:
``\st{advanced knowledge of} \kn{...}'') around the components, including degree. This goes for both before and after:
\begin{itemize}
    \item \st{Working proficiency in} \kn{developmental toolsets}.
    \item \st{Advanced knowledge of} \kn{application data and architecture infrastructure} \st{disciplines}.
    \item \sk{Manual handling} \st{tasks}.
    \item \kn{CI/CD} \st{experience}.
    \item \st{You master} \kn{English} \st{on level C1}.
    \item \st{Proficient in} \kn{Python} \st{and} \kn{English}.
    \item \st{Fluent in spoken and written} \kn{English}.
\end{itemize}
}

\guide{
7. Pay attention to expressions such as ``participation in...'', ``contributing'', and ``transfer (knowledge)''. These are usually not considered skills.
\begin{itemize}
    \item \st{Contribute to the enjoyable and collaborative work environment}.
    \item \st{Participation in the Department's regular research activities}.
    \item \st{Desire to be part of something meaningful and innovative}.
\end{itemize}
}

\guide{
8. Skills and Knowledge components that are found in not-so-straightforward places (e.g., project descriptions) are annotated as well, if they relate to the position.
}

\guide{
9. In the pattern of ``skill'' followed by some elaboration, see if it can be annotated with a skill and a knowledge tag:
\begin{itemize}
    \item \sk{Ensure food storage and preparation areas are maintained} according to \kn{Health \& Safety and Audit standards}.
\end{itemize}
}
\clearpage
\guide{
10. Occupations and positions in companies/academia should be excluded.
}
\guide{
11. If there's a knowledge/skill component in the position, we exclude it as well.
\begin{itemize}
\item Experienced Java Engineer. → completely untagged.
\end{itemize}
}
\guide{
12. Only annotate the skills that are \underline{related} to the position.\\\\
12.1. This includes skills that are specific for the position as well (e.g., skills of a ruminants professor versus math professor).\\\\
12.2 Also skills that the person for the position is expected to do in the future.\\\\
12.3 This does \underline{not} include skills, knowledge or attitudes describing only the company, the group you will join in the department, and so on. \underline{Only annotate} if it is specified or implied that the employee should possess the skill as well.
}
\guide{
13. We annotate industries and fields (that the employee will be working in) as knowledge components.}

\clearpage

\section{Type of Skills Annotated}\label{app:quali}
In both~\cref{tab:freq-skill} and~\cref{tab:freq-knowledge}, we show the top-10 skill and knowledge components that have been annotated. We split the top-10 among the data splits (i.e., train, development, and test set), and also between source splits (i.e., \bjo{}, \hou{}, \tech{}).

\section{Reproducibility}\label{app:hyper}

\begin{table}[h]
    \centering
    \resizebox{\linewidth}{!}{
    \begin{tabular}{l|r|r}
    \toprule
    \textbf{Parameter} & \textbf{Value} & \textbf{Range} \\
    \midrule
    Optimizer                           & AdamW                & \\
    $\beta_\text{1}$, $\beta_\text{2}$  & 0.9, 0.99            & \\
    Dropout                             & 0.2                  & 0.1, 0.2, 0.3\\
    Epochs                              & 20                   & \\
    Batch Size                          & 32                   & \\
    Learning Rate (LR)                  & 1e-4                 & 1e-3, 1e-4, 1e-5\\
    LR scheduler                        & Slanted triangular   & \\
    Weight decay                        & 0.01                 & \\
    Decay factor                        & 0.38                 & 0.35, 0.38, 0.5\\
    Cut fraction                        & 0.2                  & 0.1, 0.2, 0.3\\
    \bottomrule
    \end{tabular}}
    \caption{Hyperparameters of \textsc{MaChAmp}.}
    \label{tab:hyperparameters}
\end{table}

\noindent
We use the default hyperparameters in \textsc{MaChAmp}~\cite{van-der-goot-etal-2021-massive} as shown in~\cref{tab:hyperparameters}. For more details we refer to their paper. For the five random seeds we use 3477689, 4213916, 6828303, 8749520, and 9364029. All experiments with \textsc{MaChAmp} were ran on an NVIDIA\textsuperscript{\textregistered} TITAN X (Pascal) 12 GB GPU
and an Intel\textsuperscript{\textregistered} Xeon\textsuperscript{\textregistered} Silver 4214 CPU.

\begin{table*}[ht]
\centering
\resizebox{.85\linewidth}{!}{
    \begin{tabular}{l|l|ll|ll|ll}
    \toprule
    & \textbf{Evaluation} $\rightarrow$ & \multicolumn{2}{c|}{\textbf{\textsc{Skill}}} & \multicolumn{2}{c|}{\textbf{\textsc{Knowledge}}} & \multicolumn{2}{c}{\textbf{\textsc{Combined}}}\\
    \midrule
    \textbf{Src.} & $\downarrow$ \textbf{Model}, \textbf{Task} $\rightarrow$ & \textbf{STL} & \textbf{MTL} & \textbf{STL} & \textbf{MTL} & \textbf{STL (*2)} & \textbf{MTL}\\
    \midrule
    \multirow{4}{*}{\rotatebox[origin=c]{90}{\textbf{\bjo{}}}}
    & \bertb        & \std{59.55}{0.97} & \std{58.88}{1.14} & \std{50.68}{3.25} & \std{51.10}{1.67} & \std{57.46}{1.19} & \std{57.00}{0.91}\\
    & SpanBERT      & \std{59.78}{0.44} & \std{60.02}{2.15} & \std{50.65}{2.32} & \std{51.79}{2.12} & \std{57.71}{0.53} & \std{58.00}{2.07}\\
    & JobBERT       & \std{60.60}{0.81} & \std{59.76}{0.60} & \std{50.29}{1.86} & \std{47.59}{1.11} & \std{58.19}{0.49} & \std{56.75}{0.50}\\
    & JobSpanBERT   & \std{60.16}{0.61} & \std{59.44}{1.11} & \std{45.20}{2.76} & \std{47.69}{3.38} & \std{56.56}{0.49} & \std{56.58}{0.63}\\
    \midrule
    \multirow{4}{*}{\rotatebox[origin=c]{90}{\textbf{\hou{}}}}
    & \bertb        & \std{56.83}{1.29} & \std{55.89}{1.90} & \std{55.00}{1.11} & \std{54.05}{1.00} & \std{56.17}{0.92} & \std{55.20}{1.35}\\
    & SpanBERT      & \std{57.54}{1.08} & \std{57.30}{0.84} & \std{52.01}{1.72} & \std{51.48}{1.01} & \std{55.55}{1.10} & \std{55.09}{0.74}\\
    & JobBERT       & \std{59.81}{1.17} & \std{59.97}{0.85} & \std{54.94}{1.15} & \std{54.23}{2.60} & \std{58.02}{0.93} & \std{57.80}{1.50}\\
    & JobSpanBERT   & \std{59.97}{1.03} & \std{59.62}{0.74} & \std{55.66}{1.51} & \std{53.10}{1.27} & \std{58.37}{1.07} & \std{57.14}{0.56}\\
    \midrule
    \multirow{4}{*}{\rotatebox[origin=c]{90}{\textbf{\tech{}}}}
    & \bertb        & \std{59.05}{0.71} & \std{58.34}{0.75} & \std{64.08}{1.04} & \std{63.77}{1.18} & \std{62.10}{0.67} & \std{61.65}{0.62}\\
    & SpanBERT      & \std{58.39}{0.46} & \std{58.61}{1.14} & \std{62.68}{0.60} & \std{63.40}{0.93} & \std{61.02}{0.35} & \std{61.56}{0.81}\\
    & JobBERT       & \std{59.81}{0.75} & \std{59.36}{0.90} & \std{64.57}{0.42} & \std{63.15}{0.94} & \std{62.69}{0.40} & \std{61.67}{0.90}\\
    & JobSpanBERT   & \std{60.09}{1.43} & \std{59.48}{0.61} & \std{63.40}{1.51} & \std{63.23}{0.64} & \std{62.09}{0.85} & \std{61.80}{0.54}\\
    \midrule
    \midrule
    \rowcolor{Gray}
    & \bertb        & \std{58.45}{0.68} & \std{57.67}{1.01} & \std{60.44}{0.58} & \std{59.98}{0.75} & \std{59.35}{0.46} & \std{58.72}{0.48}\\
    \rowcolor{Gray}
    & SpanBERT      & \std{58.53}{0.33} & \std{58.60}{0.83} & \std{58.89}{0.49} & \std{59.21}{0.78} & \std{58.69}{0.36} & \std{58.88}{0.64}\\
    \rowcolor{Gray}
    & JobBERT       & \std{60.05}{0.70} & \std{59.69}{0.62} & \textbf{\std{60.66}{0.43}}\textsuperscript{*} & \std{59.15}{1.07} & \textbf{\std{60.32}{0.39}}\textsuperscript{*} & \std{59.44}{0.81}\\
    \rowcolor{Gray}
    \multirow{-4}{*}{\rotatebox[origin=c]{90}{\textbf{\textsc{Average}}}}
    & JobSpanBERT   & \textbf{\std{60.07}{0.30}}\textsuperscript{$\dagger$} & \std{59.51}{0.68} & \std{59.47}{1.31} & \std{59.04}{0.65} & \std{59.79}{0.53} & \std{59.29}{0.43}\\
    \midrule
    \midrule
    \rowcolor{LightCyan}
    & \bertb        & \std{54.34}{0.74} & \std{54.20}{0.68} & \std{62.43}{0.41} & \std{61.66}{0.83} & \std{58.16}{0.47} & \std{57.73}{0.66} \\
    \rowcolor{LightCyan}
    & JobBERT       & \std{56.11}{0.49} & \std{55.46}{0.75} & \textbf{\std{63.88}{0.28}}\textsuperscript{*} & \std{63.35}{0.30} & \textbf{\std{59.73}{0.38}}\textsuperscript{$\dagger$} & \std{59.18}{0.37}\\
    \rowcolor{LightCyan}
    \multirow{-3}{*}{\rotatebox[origin=c]{90}{\textbf{\textsc{Test}}}}
    & JobSpanBERT   & \textbf{\std{56.64}{0.83}}\textsuperscript{*} & \std{56.27}{0.55} & \std{61.06}{0.99} & \std{61.87}{0.55} & \std{58.72}{0.69} & \std{58.90}{0.48}\\
    \bottomrule
    \end{tabular}
    }
\caption{\textbf{Performance of Models.} We test the models on \textbf{skills}, \textbf{\textsc{Knowledge}}, and \textbf{\textsc{Combined}} (MTL). We report the span-F1 and standard deviation of runs on five random seeds on the \emph{development set} (\textbf{\textsc{Average}}, in gray). Results on the \emph{test set} are below in the \textbf{\textsc{Test}} rows (in cyan). \textbf{STL} indicates single-task learning and \textbf{MTL} indicates the multi-task model. 
\textbf{Bold} numbers indicate best performing model in that experiment. A ($\dagger$) means that it is stochastically dominant over \emph{all} the other models. (*) denotes \emph{almost stochastic dominance }($\epsilon_\text{min} <$ 0.5) over---at minimum---one other model.}
    \label{tab:results}
\end{table*}

\begin{table*}[t]
\centering
\resizebox{.85\linewidth}{!}{
    \begin{tabular}{l|l|rr|rr|rr}
    \toprule
& \textbf{Evaluation} $\rightarrow$ & \multicolumn{2}{c|}{\textbf{\textsc{Skill}}} & \multicolumn{2}{c|}{\textbf{\textsc{Knowledge}}} & \multicolumn{2}{c}{\textbf{\textsc{Multi}}}\\
\midrule
\textbf{Src.} & $\downarrow$ \textbf{Model} & \textbf{Precision} & \textbf{Recall} & \textbf{Precision} & \textbf{Recall} & \textbf{Precision} & \textbf{Recall}\\
\midrule
\multirow{4}{*}{\rotatebox[origin=c]{90}{\textbf{\bjo{}}}}
    & \bertb        & \std{57.09}{1.70} & \std{62.27}{1.28}       & \std{43.95}{4.17} & \std{60.00}{1.65}               & \std{52.63}{1.32} & \std{62.19}{0.87} \\
    & SpanBERT      & \std{58.28}{0.59} & \std{61.36}{0.68}       & \std{45.80}{2.89} & \std{56.82}{3.39}               & \std{54.02}{1.81} & \std{62.63}{2.60} \\
    & JobBERT       & \std{57.90}{1.25} & \std{63.59}{0.99}       & \std{43.45}{1.98} & \std{59.84}{3.44}               & \std{51.13}{0.48} & \std{63.74}{0.79} \\
    & JobSpanBERT   & \std{58.39}{1.03} & \std{62.09}{1.85}       & \std{38.55}{3.12} & \std{54.76}{3.18}               & \std{52.22}{0.35} & \std{61.75}{1.22} \\
    \midrule
\multirow{4}{*}{\rotatebox[origin=c]{90}{\textbf{\hou{}}}}
    & \bertb        & \std{55.95}{2.46} & \std{57.79}{0.67}      & \std{52.84}{0.65} & \std{57.42}{2.76}                & \std{51.65}{1.11} & \std{59.28}{2.07} \\
    & SpanBERT      & \std{56.70}{1.59} & \std{58.44}{1.16}      & \std{49.87}{2.57} & \std{54.49}{3.09}                & \std{52.27}{0.64} & \std{58.25}{1.50} \\
    & JobBERT       & \std{58.16}{1.30} & \std{61.56}{1.53}      & \std{51.18}{2.18} & \std{59.37}{1.34}                & \std{53.72}{1.57} & \std{62.56}{1.47} \\
    & JobSpanBERT   & \std{59.04}{0.85} & \std{60.99}{2.58}      & \std{51.36}{2.70} & \std{60.84}{1.19}                & \std{53.91}{0.77} & \std{60.79}{0.54} \\
     \midrule
\multirow{4}{*}{\rotatebox[origin=c]{90}{\textbf{\tech{}}}}
    & \bertb        & \std{58.28}{1.30} & \std{59.89}{1.39}      & \std{60.79}{1.89} & \std{67.79}{1.20}                & \std{58.19}{1.12} & \std{65.55}{0.75} \\
    & SpanBERT      & \std{58.62}{0.32} & \std{58.16}{0.76}      & \std{59.43}{1.21} & \std{66.35}{1.18}                & \std{58.34}{0.97} & \std{65.17}{1.41} \\
    & JobBERT       & \std{58.81}{1.38} & \std{60.88}{1.51}      & \std{61.38}{1.11} & \std{68.14}{1.36}                & \std{57.69}{0.93} & \std{66.25}{0.90} \\
    & JobSpanBERT   & \std{59.86}{3.07} & \std{60.40}{0.68}      & \std{59.78}{2.43} & \std{67.57}{1.97}                & \std{58.26}{0.82} & \std{65.82}{0.91} \\
    \midrule
     \midrule
\multirow{4}{*}{\rotatebox[origin=c]{90}{\textbf{\textsc{Average}}}}
    & \bertb        & \std{57.11}{1.65} & \std{59.90}{0.95}      & \std{56.86}{1.33} & \std{64.54}{1.31}                & \std{55.02}{0.85} & \std{62.98}{0.93} \\
    & SpanBERT      & \std{57.85}{0.65} & \std{59.23}{0.52}      & \std{55.65}{1.09} & \std{62.58}{1.56}                & \std{55.61}{0.61} & \std{62.58}{1.25} \\
    & JobBERT       & \std{58.29}{1.08} & \std{61.94}{1.16}      & \std{56.73}{1.41} & \std{65.22}{1.03}                & \std{55.03}{0.84} & \std{64.62}{0.77} \\
    & JobSpanBERT   & \std{59.11}{1.59} & \std{61.12}{1.49}      & \std{55.11}{2.41} & \std{64.66}{1.38}                & \std{55.64}{0.56} & \std{63.46}{0.69} \\
     \midrule
     \midrule
     \multirow{3}{*}{\rotatebox[origin=c]{90}{\textbf{\textsc{Test}}}}
    & \bertb        & \std{56.02}{1.50}  & \std{52.79}{1.18}      & \std{59.09}{0.85} & \std{66.20}{1.69}                & \std{55.82}{1.03} & \std{59.79}{0.87} \\
    & JobBERT       & \std{55.94}{1.19} & \std{56.29}{0.49}      & \std{60.03}{1.13} & \std{68.30}{1.46}                & \std{55.87}{0.29} & \std{62.89}{0.56}\\
    & JobSpanBERT   & \std{57.57}{1.24} & \std{55.77}{1.65}      & \std{57.83}{1.03} & \std{64.71}{2.10}                & \std{57.06}{0.74} & \std{60.89}{0.42} \\
    \bottomrule
    \end{tabular}
    }
\caption{\textbf{Precision and Recall of Models}. We test the models on \emph{skills}, \emph{knowledge}, and \emph{multi-task} setting. We report the average precision, recall and standard deviation of runs on five random seeds on the \emph{development set} (\textbf{\textsc{Average}}). Results on the \emph{test set} are below in the \textbf{\textsc{Test}} rows.}
    \label{tab:results-precision-recall}
\end{table*}

\begin{table*}[ht]
    \centering
\resizebox{.86\linewidth}{!}{
    \begin{tabular}{l|rr|rr|rr}
    \toprule
    \textbf{Source} $\rightarrow$& \multicolumn{2}{c|}{\textbf{\bjo{}}} & \multicolumn{2}{c|}{\textbf{\hou{}}} & \multicolumn{2}{c}{\textbf{\tech{}}}\\
    \midrule
    $\downarrow$ \textbf{Model} & \textsc{\textbf{Skills}} (\#) & \textbf{\textsc{Knowledge}} (\#) & \textsc{\textbf{Skills}} (\#) & \textbf{\textsc{Knowledge}} (\#) & \textsc{\textbf{Skills}} (\#) & \textbf{\textsc{Knowledge}} (\#)\\
\midrule
    \textsc{Annotations}      & 4.16 (634)             & 2.03 (242)             & 3.81 (637)             & 1.91 (350)             & 3.92 (459) & 1.69 (834) \\
     \bertb                    & \std{4.42}{0.11} (628) & \std{2.17}{0.06} (307) & \std{3.89}{0.11} (615) & \std{1.98}{0.04} (461) & \std{4.43}{0.06} (449) & \std{1.75}{0.02} (885)\\
     SpanBERT                  & \std{4.50}{0.04} (621) & \std{2.14}{0.03} (298) & \std{3.92}{0.04} (597) & \std{2.03}{0.03} (441) & \std{4.33}{0.06} (444) & \std{1.76}{0.03} (869)\\
     JobBERT                   & \std{4.38}{0.11} (670) & \std{2.10}{0.06} (313) & \std{3.97}{0.08} (650) & \std{1.99}{0.04} (470) & \std{4.42}{0.10} (479) & \std{1.72}{0.03} (932)\\
     JobSpanBERT               & \std{4.51}{0.09} (629) & \std{2.08}{0.05} (313) & \std{3.95}{0.11} (623) & \std{2.01}{0.06} (452) & \std{4.48}{0.12} (439) & \std{1.71}{0.03} (875)\\
     Longformer                & \std{4.45}{0.14} (653) & \std{2.22}{0.04} (298) & \std{3.90}{0.17} (639) & \std{1.97}{0.03} (483) & \std{4.40}{0.10} (472) & \std{1.80}{0.05} (864)\\
\bottomrule
    \end{tabular}}
    \caption{\textbf{Average Length of Predictions of Single Models.} We show the average length of the predictions versus the length of our annotated skills and knowledge components on the \emph{test set} and the total number of predicted skills and knowledge tags in each respective split (\#).}
    \label{tab:avglength}
\end{table*}

\section{Exact Number of Performance}\label{app:source-perf}
In~\cref{tab:results}, we show the exact numbers of the plot indicated in~\cref{fig:results}. In addition, we also show the results of each respective split.

For the STL models, we observe differences in performances over the sources which is particularly pronounced for knowledge components: The \textsc{Tech} source is the easiest to process (and has most SKCs), while SKCs identification performance is the lowest for \textsc{Big}. This might be due to the broad nature of this source. 

In the exact results table (\cref{tab:results}) we add a ($\dagger$) next to the highest span-F1 if the model is truly stochastically dominant ($\epsilon_\text{min} = $ 0.0) over \emph{all} the other models. (*) denotes that the best model achieved \emph{almost stochastic dominance} ($\epsilon_\text{min} <$ 0.5) over---at minimum---one other model (e.g., in \textsc{Test} rows w.r.t \textsc{Combined}: MTL-JobBERT $\succeq$ MTL-JobSpanBERT with $\epsilon_\text{min} =$ 0.06) and stochastically dominant over the rest.

In~\cref{tab:results-precision-recall}, we report the precision and recall of the models, \textsc{Skill} and \textsc{Knowledge} show the precision and recall of the STL models. \textsc{Multi} shows the precision and recall of the MTL models.

Last, in \cref{tab:avglength}, we show the exact numbers of the length of predictions~\cref{fig:avglength}. We also add the number of predicted \textsc{Skill} and \textsc{Knowledge} Overall, JobBERT and JobSpanBERT predict more skills in general than the other models. This is also the case for knowledge components. We hypothesize that this might be due to the BERT models now being more specialized towards the JP domain and recognizing more SKCs.

\begin{figure*}
    \centering
    \includegraphics[width=.32\linewidth]{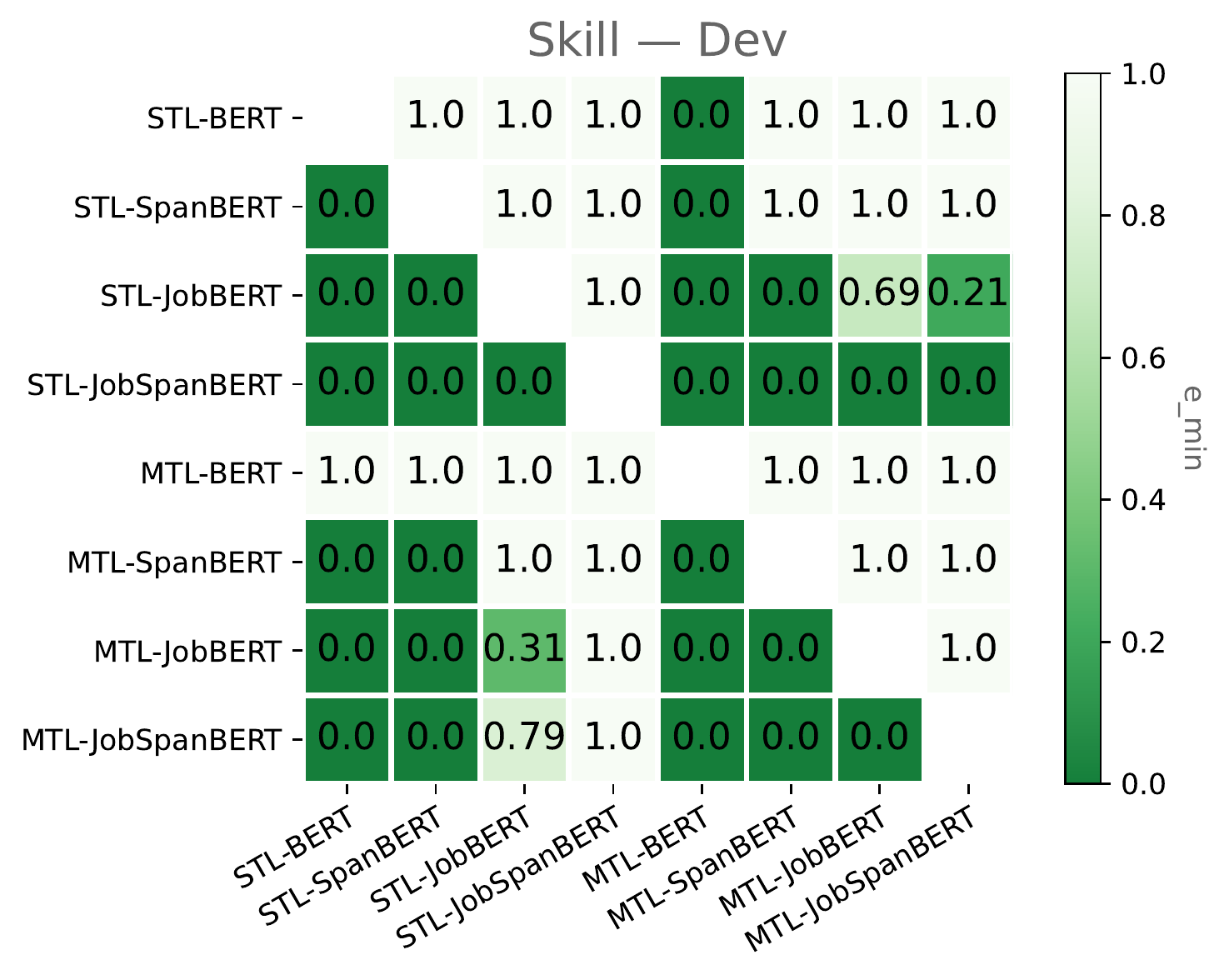}
    \includegraphics[width=.32\linewidth]{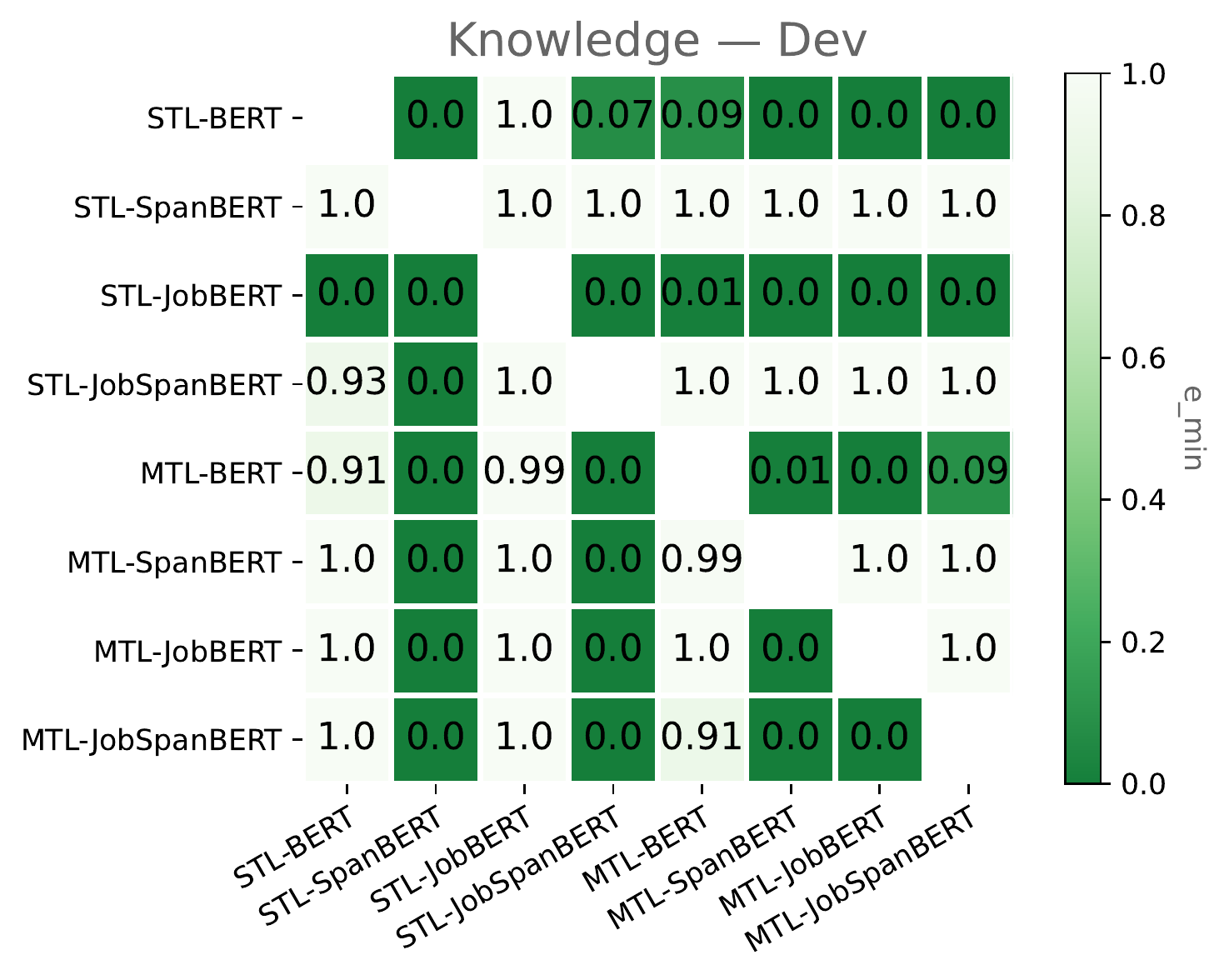}
    \includegraphics[width=.32\linewidth]{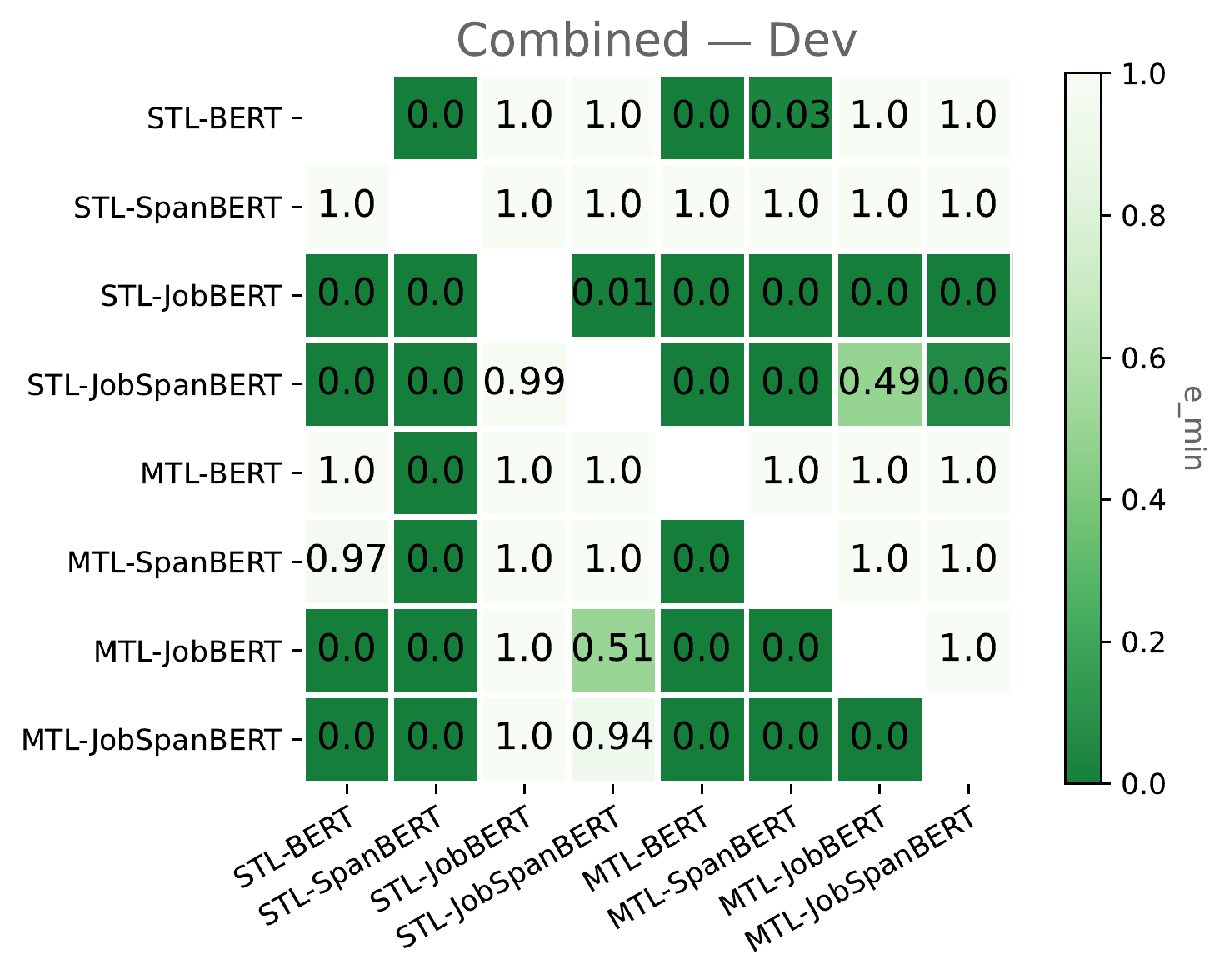}
    \caption{\textbf{Almost Stochastic Order Scores of the Development Set.}
    ASO scores expressed in $\epsilon_\text{min}$.
    The significance level $\alpha =$ 0.05 is adjusted accordingly by using the Bonferroni correction~\citep{bonferroni1936teoria}. Read from row to column: E.g., STL-JobBERT (row) is stochastically dominant over STL-\bertb{} (column) with $\epsilon_\text{min}$ of 0.00.}
    \label{fig:significance-dev}
\end{figure*}

\section{Significance Testing}\label{app:sign-per-source}
Recently, the ASO test~\cite{dror2019deep}\footnote{Implementation of \citet{dror2019deep} can be found at~\url{https://github.com/Kaleidophon/deep-significance}~\cite{dennis_ulmer_2021_4638709}} has been proposed to test statistical significance for deep neural networks over multiple runs.
Generally, the ASO test determines whether a stochastic order~\citep{reimers2018comparing} exists between two models or algorithms based on their respective sets of evaluation scores. Given the single model scores over multiple random seeds of two algorithms $\mathcal{A}$ and $\mathcal{B}$, the method computes a test-specific value ($\epsilon_\text{min}$) that indicates how far algorithm $\mathcal{A}$ is from being significantly better than algorithm $\mathcal{B}$. When distance $\epsilon_\text{min} = 0.0$, one can claim that $\mathcal{A}$ stochastically dominant over $\mathcal{B}$ with a predefined significance level. When $\epsilon_\text{min} < 0.5$ one can say $\mathcal{A} \succeq \mathcal{B}$. On the contrary, when we have $\epsilon_\text{min} = 1.0$, this means $\mathcal{B} \succeq \mathcal{A}$. For $\epsilon_\text{min} = 0.5$, no order can be determined. We took 0.05 for the predefined significance level. In \cref{fig:significance-dev}, we show the ASO scores on the development set.

\clearpage
\begin{landscape}
\begin{table}
\centering
    \resizebox{1.3\textwidth}{!}{
    \begin{tabular}{l|lll}
    \toprule
& \multicolumn{3}{c}{\textbf{Skills}}\\
\midrule
\textbf{Src.} & \textbf{Train} & \textbf{Development} & \textbf{Test}\\
\midrule
\multirow{10}{*}{\rotatebox[origin=l]{90}{\textbf{\bjo{}}}}
    & enthusiastic          & ambitious                              & customer service                                              \\
    & flexible              & proactive                              & communicator                                                  \\
    & team player           & work independently                     & flexible                                                      \\
    & friendly              & attention to detail                    & attention to detail                                           \\
    & attention to detail   & motivated                              & ambitious                                                     \\
    & communicator          & reliable                               & design and refine every touchpoint of the customer journey    \\
    & passionate            & flexible                               & enable inclusion                                              \\
    & communication         & willingness to learn                   & communicate effectively                                       \\
    & confident             & self-motivated                         & interpersonal skills                                          \\
    & flexible approach     & work as part of an established team    & proactive                                                     \\
    \midrule
\multirow{10}{*}{\rotatebox[origin=l]{90}{\textbf{\hou{}}}}
    & communication skills  & structured              & teaching              \\
    & motivated             & teaching                & research              \\
    & structured            & communication skills    & communication skills  \\
    & proactive             & project management      & outgoing              \\
    & analytical            & drive                   & flexible              \\
    & communication         & problem solving         & energetic             \\
    & self-driven           & communication           & responsible           \\
    & team player           & visit customers         & enthusiastic          \\
    & teaching              & curious                 & team player           \\
    & curious               & work independently      & communication         \\
    \midrule
\multirow{10}{*}{\rotatebox[origin=l]{90}{\textbf{\tech{}}}}
    & communication skills                                                      & hands-on               & solving business problems                             \\
    & passionate                                                                & communication skills   & apply your depth of knowledge and expertise           \\
    & apply your depth of knowledge and expertise                               & leadership             & partner continuously with your many stakeholders      \\
    & partner continuously with your many stakeholders                          & passionate             & achieve organizational goals                          \\
    & solving business problems through innovation and engineering practices    & open-minded            & building an innovative culture                        \\
    & work in large collaborative teams                                         & code reviews           & stay focused on common goals                          \\
    & hands-on                                                                  & independent            & work in large collaborative teams                     \\
    & building an innovative culture                                            & software development   & design                                                \\
    & team player                                                               & pioneer new approaches & development                                           \\
    & develop                                                                   & analytical skills      & communicate                                           \\
    \bottomrule
    \end{tabular}}
    \caption{\textbf{Most Frequent Skills in the Data.} Top--10 skill components in our data in terms of frequency.}
    \label{tab:freq-skill}
\end{table}
\end{landscape}

\clearpage

\begin{table*}
\centering
\resizebox{\linewidth}{!}{
    \begin{tabular}{l|lll}
    \toprule
& \multicolumn{3}{c}{\textbf{Knowledge}}\\
\midrule
\textbf{Src.} & \textbf{Train} & \textbf{Development} & \textbf{Test}\\
\midrule
\multirow{10}{*}{\rotatebox[origin=l]{90}{\textbf{\bjo{}}}}
 & english              & full uk driving licence                                & strategic planning    \\
 & driving license      & sap energy assessments                                 & english               \\
 & excel                & right to work in the uk                                & cscs card             \\
 & cscs card            & sen                                                    & pms                   \\
 & maths                & acca/aca                                               & reservation systems   \\
 & ppc                  & professional kitchen                                   & keynote               \\
 & service design       & cra calculations                                       & illustrator           \\
 & uk/emea policies     & email marketing                                        & aba                   \\
 & bachelor’s degree    & qualitative and quantitative social research methods   & sen                   \\
 & computer science     & care setting                                           & full driving license  \\
    \midrule
\multirow{10}{*}{\rotatebox[origin=l]{90}{\textbf{\hou{}}}}
 & english              & english                                    & english         \\
 & engineering          & supply chain                               & danish          \\
 & computer science     & project management                         & business        \\
 & product management   & powders                                    & java            \\
 & python               & machine learning                           & marketing       \\
 & finance              & phd degree                                 & plm             \\
 & project management   & muscle models with learning and adaptation & production      \\
 & agile                & walking robots                             & supply chain    \\
 & danish               & model rules                                & economics       \\
 & javascript           & capacity development                       & excel           \\
    \midrule
\multirow{10}{*}{\rotatebox[origin=l]{90}{\textbf{\tech{}}}}
 & javascript           & java                 & java          \\
 & python               & javascript           & python        \\
 & java                 & aws                  & .net          \\
 & agile                & docker               & financial services \\
 & financial services   & node.js              & c\#           \\
 & node.js              & typescript           & javascript        \\
 & english              & react                & cloud             \\
 & kubernetes           & linux                & english           \\
 & cloud                & amazon-web-services  & reactjs           \\
 & docker               & devops               & automation        \\
    \bottomrule
    \end{tabular}}
    \caption{\textbf{Most Frequent Knowledge in the Data.} Top--10 knowledge components in our data in terms of frequency.}
    \label{tab:freq-knowledge}
\end{table*}

\end{document}